\documentclass[10pt,twocolumn,letterpaper]{article}

\usepackage[pagenumbers]{wacv}      % To produce the REVIEW version for the algorithms track
\definecolor{wacvblue}{rgb}{0.21,0.49,0.74}
\usepackage[pagebackref,breaklinks,colorlinks,allcolors=wacvblue]{hyperref}
\usepackage[table]{xcolor}
\def\wacvPaperID{736} % *** Enter the WACV Paper ID here
\def\confName{WACV}
\def\confYear{2027}

\title{SAGE: Semantic-Augmented Geometric Experts for Point Cloud Registration}

\author{Tomoyasu Shimada \qquad Kazuhiko Murasaki \qquad Shoichiro Takeda \qquad Shogo Sato\\
Riku Inoue \qquad  Toshihiko Nishimura \qquad  Ryuichi Tanida\\
NTT, Inc.}
\begin{document}
\maketitle
\begin{abstract}
Point cloud registration is an essential task for scene reconstruction, requiring the alignment of partially overlapping point clouds. Under low-overlap conditions, this task remains challenging because existing geometry-only methods cannot distinguish geometrically similar but semantically distinct structures (e.g., TV corners vs. desk corners and chair legs vs. table legs), resulting in incorrect correspondences. To address this geometric ambiguity, we present \textbf{S}emantic-\textbf{A}ugmented \textbf{G}eometric \textbf{E}xperts (\textbf{SAGE}), which integrates frozen pre-trained point cloud semantic representations into the registration pipeline without requiring additional sensors or fine-tuning the semantic encoder. To effectively leverage these semantic distinctions, SAGE employs a Semantic-Guided Mixture of Experts (SG-MoE) architecture. Through semantic-aware routing, SG-MoE enables each expert to specialize in distinct semantic-geometric patterns, reducing mismatches that confound geometry-only methods. With a structure-aware estimator, SAGE achieves the highest registration recall on 3DLoMatch while remaining competitive on 3DMatch. When evaluated with the same number of refinement iterations, SAGE further achieves the highest inlier ratio and registration recall on both benchmarks. These results indicate that frozen semantic representations provide effective cues for resolving geometric ambiguity in low-overlap registration.
\end{abstract}
\section{Introduction}
\label{sec:intro}

\begin{figure}[t]
\centering
\includegraphics[width=.85\linewidth]{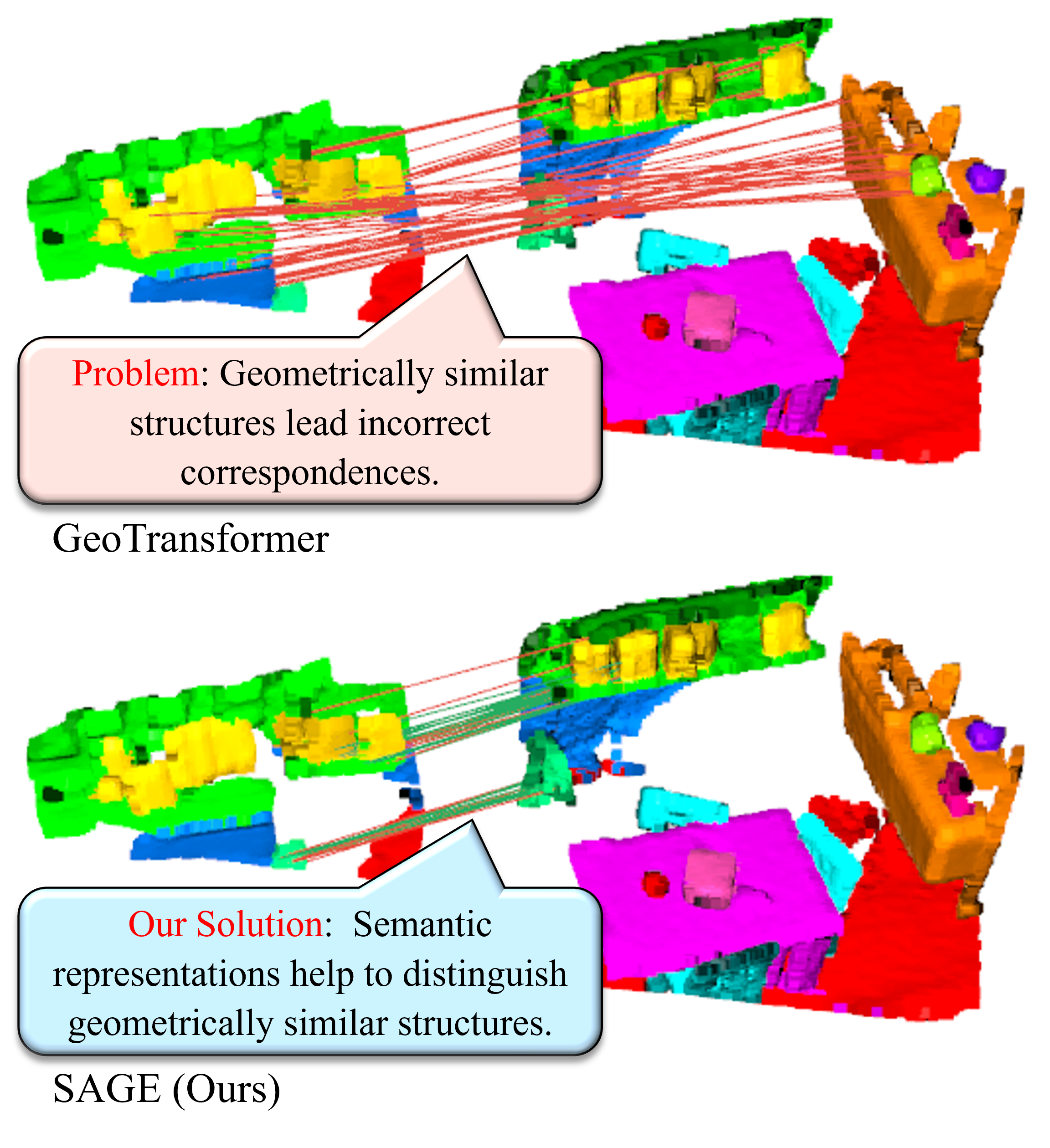}
\vspace{-8pt}
\caption{
\textbf{Limitation of geometric features in low-overlap point cloud registration.}
Point clouds are pseudo-colored by semantic regions for visualization.
\textbf{Top:} GeoTransformer~\cite{qin2022geotransformer}, relying solely on geometric features, is overwhelmed by dominant incorrect correspondences (red lines) due to similar geometric structures, \textit{failing to establish true inliers (no green lines visible)}.
\textbf{Bottom:} SAGE (Ours) successfully disambiguates geometrically similar but semantically distinct structures.
Green and red lines indicate inlier and outlier correspondences (threshold: 0.1\,m).
}
%\vspace{-16pt}
\label{fig:intro}
\end{figure}

Point cloud registration is an essential task for 3D perception in robotics, autonomous driving, and augmented reality~\cite{huang2021comprehensive}.
Despite progress, achieving robust registration under challenging conditions with low overlap remains an open problem. 
Current registration methods~\cite{huang2021predator,qin2022geotransformer,yew2022regtr,yu2023roitr,huang2025psreg} generally follow a correspondence-based pipeline.
This pipeline first establishes point-to-point correspondences between two point clouds using deep neural networks, and then estimates the transformation from these correspondences.
Thus, this pipeline's success depends on the quality of the correspondences.
However, geometrically similar structures such as TV edges, desk edges, and furniture edges produce similar geometric features even though they differ semantically.
In low-overlap scenarios where spatial context is severely limited, correspondences relying solely on geometric features become unreliable because there are insufficient points surrounding the target point to distinguish between semantically different features.

Fig.~\ref{fig:intro} illustrates this fundamental limitation: widely adopted methods such as GeoTransformer~\cite{qin2022geotransformer} can produce entirely incorrect correspondences when faced with geometrically similar but semantically distinct structures in low-overlap scenarios.

Indeed, geometry-based methods~\cite{yu2021cofinet,huang2021predator,yew2022regtr,yu2023roitr} have such limitations.
Even recent Mixture of Experts (MoE) methods like PSReg~\cite{huang2025psreg}, which use geometric feature-based routing to activate geometry-specialized experts, depend only on geometric representations.
Although post-processing methods such as DecisionPCR~\cite{zhang2025decisionpcr} can boost results through iterative refinement, they still operate within the geometric domain, limited by the quality ceiling of geometry-based correspondence establishment.
As one possible approach to overcome this limitation, several methods use semantic information for registration.
However, they often require RGB-D sensors or calibrated cameras, while others rely on domain-specific models that generalize poorly across datasets~\cite{xu2025s2reg,fung2024semreg,mu2024colorpcr,qin2024sarnet}.

To overcome the fundamental limitations of relying solely on geometric features, we introduce \textbf{S}emantic-\textbf{A}ugmented \textbf{G}eometric \textbf{E}xperts (\textbf{SAGE}).
SAGE integrates frozen pre-trained semantic representations into point cloud registration through a semantic-guided mixture-of-experts architecture that jointly exploits geometric and semantic features.
This design requires neither additional sensors nor fine-tuning of the semantic encoder.
First, SAGE employs the frozen semantic feature extractor of the recent pre-trained model Sonata~\cite{wu2025sonata} to extract semantic features of the input source and target point clouds.
Then, SAGE jointly encodes the semantic features with input point clouds of the source and target to extract their corresponding semantic-geometric features.
After this feature extraction, SAGE refines the semantic-geometric features via our proposed Semantic-Guided MoE (SG-MoE).

Our main technical contribution is the design of SG-MoE.
If Standard MoE~\cite{shazeer2017outrageously,fedus2022switch} is used for feature refinement, as in PSReg~\cite{huang2025psreg}, geometric knowledge becomes dominant and semantic knowledge is neglected in the semantic-geometric features because the loss functions commonly adopted in registration tasks primarily optimize geometric fidelity.
To prevent this domination, we design the expert routing in SG-MoE to be based solely on semantic features extracted from the frozen Sonata.
This design enables each expert to refine the semantic-geometric features by explicitly considering semantic knowledge.
As a result, SG-MoE performs the feature refinement process while preserving both semantic and geometric information.

Finally, SAGE estimates the transformation between the source and target point clouds using their refined semantic-geometric features following the superpoint matching in GeoTransformer~\cite{qin2022geotransformer}.
As demonstrated in Fig.~\ref{fig:intro}, while geometry-only methods like GeoTransformer fail to distinguish geometrically similar structures (producing incorrect correspondences shown in red), our SAGE successfully distinguishes them by leveraging semantic knowledge, achieving accurate registration even in low-overlap scenarios.
Our contributions are:
\begin{itemize}
    \item We propose SAGE, which integrates frozen pre-trained semantic representations into point cloud registration through a two-stage feature extraction without additional sensors or retraining, overcoming the limitations of purely geometric features.
    \item We design Semantic-Guided MoE (SG-MoE), where routing is determined solely by semantic features while experts process semantic-geometric features, enabling the distinction of geometrically similar but semantically different structures.
    \item Evaluation on 3DMatch~\cite{zeng20173dmatch}, 3DLoMatch~\cite{huang2021predator}, and KITTI~\cite{geiger2012kitti}\footnote{Detailed KITTI experiments and results are provided in Appendix D.1.} demonstrates that SAGE achieves the highest registration recall on 3DLoMatch among single-pass methods with a structure-aware estimator~\cite{qin2022geotransformer}, the best feature matching recall across all sample counts and the best inlier ratio with 1000 or fewer samples under RANSAC, and the lowest translation error on KITTI, while performing on par with iterative methods~\cite{yu2023peal,huang2025psreg}.

\end{itemize}
\section{Related Work}
\label{sec:related}

\vspace{3pt}\noindent\textbf{Classical point cloud registration.}
Traditional methods laid the foundation for automatic point cloud alignment through coarse-to-fine pipelines~\cite{huang2021comprehensive}.
These methods rely on handcrafted geometric descriptors~\cite{johnson1999using,rusu2009fpfh,tombari2010shot} to encode local geometry,
establish correspondences using RANdom SAmple Consensus (RANSAC) variants~\cite{fischler1981random,aiger2008pcs,barath2018graph,choi2015robust}, and refine alignment through Iterative Closest Point (ICP)~\cite{besl1992method,chen1992object,rusinkiewicz2001efficient,pottmann2006geometry}.
These methods successfully automated fundamental registration tasks and introduced the coarse-to-fine pipeline that remains widely adopted. 
Unfortunately, because handcrafted descriptors are not learned from data, these methods are sensitive to noise and partial observations, resulting in limited robustness.

\begin{figure*}[t]
\centering
\vspace{-12pt}
\includegraphics[width=\linewidth]{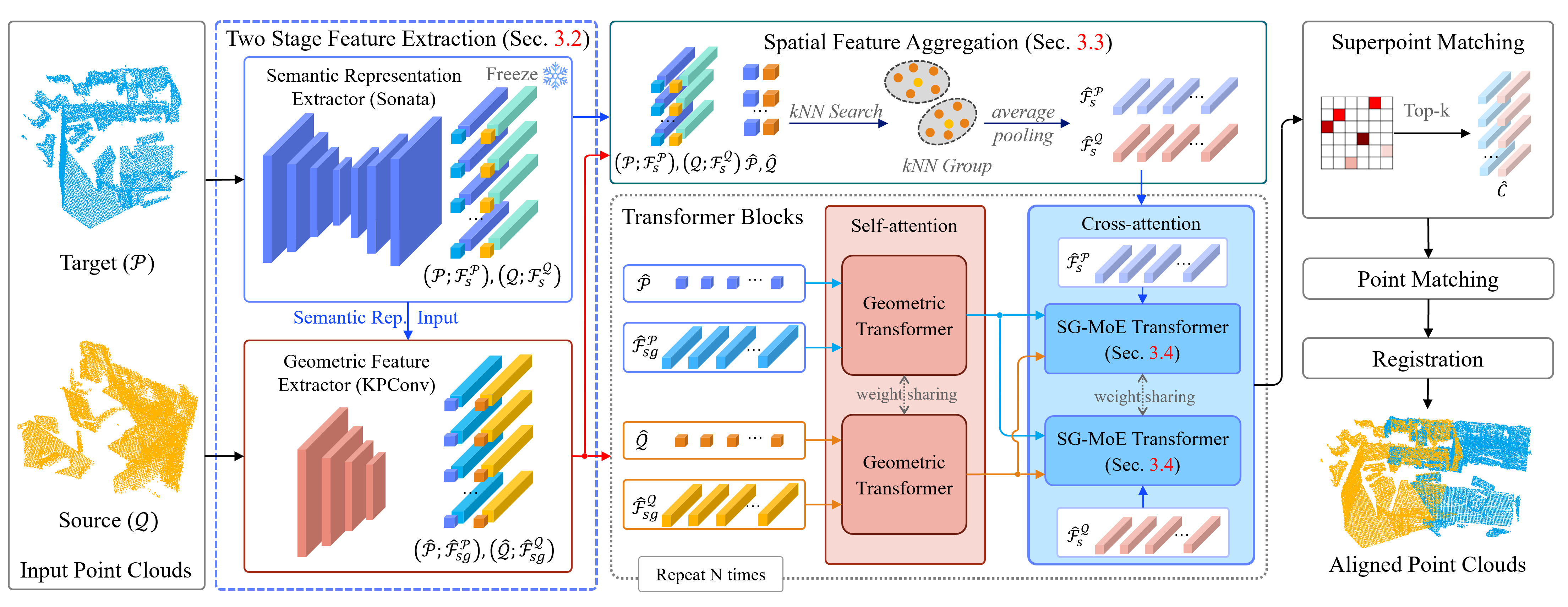}
\vspace{-12pt}
\caption{\textbf{Overall pipeline of SAGE.} 
Given two input point clouds, the \textit{Two-Stage Feature Extraction} (Sec.~\ref{subsec:feature_extraction}) first applies a frozen Sonata encoder to extract point-level semantic features from each point cloud, and then feeds these semantic features together with the raw point coordinates into the KPConv encoder to produce semantic-geometric features.
The \textit{Spatial Feature Aggregation} (Sec.~\ref{subsec:aggregation}) then maps both the semantic and semantic-geometric features to the superpoint level via nearest-neighbor search and average pooling.
The aggregated features are refined through $N$ alternating iterations of Geometric Transformer layers (red box, self-attention) and \textit{Semantic-Guided Mixture of Experts} (Sec.~\ref{subsec:sg_moe}) layers (blue box, cross-attention), where semantic features guide expert routing to enable semantic-aware specialization.
Finally, the framework adopts GeoTransformer's coarse-to-fine matching for correspondence estimation and registration.}
\label{fig:pipeline}
\end{figure*}

\vspace{3pt}\noindent\textbf{Learning-based point cloud registration.}
Deep learning addressed the discriminative power limitations of handcrafted descriptors by introducing learnable feature representations.
Pioneering works such as 3DMatch~\cite{zeng20173dmatch} for RGB-D scenes and FCGF~\cite{choy2019fully} with sparse convolutions demonstrated improvements in both feature discriminability and robustness to noise and partial observations by learning local geometric features from large-scale datasets~\cite{deng2018ppf,yang2018foldingnet,aoki2019pointnetlk,bai2020d3feat}.
However, a new challenge emerged in low-overlap scenarios where limited spatial context makes correspondence establishment particularly difficult.
Unlike partial observations (which concern the quality and completeness of individual point clouds), low-overlap refers to the limited shared observable region between two point clouds.
Even when each scan is individually complete, establishing reliable correspondences becomes difficult when the common area is small, as there is insufficient spatial context to disambiguate similar structures.
Despite improved feature quality, methods relying on local geometric features alone struggle in such scenarios~\cite{huang2021comprehensive}.
This is because they cannot leverage global scene structure to resolve ambiguities when similar local geometric patterns appear at multiple locations.

\vspace{3pt}\noindent\textbf{Global context modeling with transformers.}
Transformers addressed the locality limitation by enabling global context modeling through self-attention mechanisms~\cite{wang2019deep,yu2021cofinet,huang2021predator}.
GeoTransformer~\cite{qin2022geotransformer} pioneered this direction with geometric transformers and superpoint matching.
Subsequent work explored end-to-end correspondence prediction~\cite{yew2022regtr}, hierarchical modeling, diffusion-based approaches, multi-modal fusion, and various architectural improvements~\cite{huang2021predator,yu2023peal,chen2024full,she2024pointdifformer,she2024posdiffnet,an2024diffusion,mu2024colorpcr,shimada2024memory,wang2025point,zhang2025decisionpcr}.
By incorporating global geometric relationships, these methods improved performance in low-overlap scenarios where spatial context is limited.
However, we observe that purely geometric features struggle to disambiguate structures with similar geometric patterns but different semantic meanings.

\vspace{3pt}\noindent\textbf{Representation learning beyond geometry.}
Recent advances in 3D representation learning~\cite{wu2024mitigating,wu2023masked,wang2025unipre3d} have demonstrated the potential to learn semantic representations beyond pure geometry, offering a promising direction for addressing geometric ambiguity.
Methods such as Point-BERT~\cite{yu2022point}, masked autoencoders, and auto-regressive methods~\cite{pang2022masked,chen2023pointgpt} capture semantic patterns through pre-training with masked point modeling.
Notably, Sonata~\cite{wu2025sonata} learns robust semantic representations by explicitly avoiding geometric shortcuts during training using a large dataset.
This makes it possible to semantically distinguish between objects that have similar geometric characteristics, such as walls and floors.

\vspace{3pt}\noindent\textbf{Semantic-enhanced registration.}
Several methods leverage semantic information for registration.
S2Reg~\cite{xu2025s2reg} acquires semantics via pre-trained 3D models or 2D-3D projection from images, while ML-SemReg~\cite{yan2024ml} relies on semantic labels produced by external semantic segmentation models.
ColorPCR~\cite{mu2024colorpcr}, SemReg~\cite{fung2024semreg}, and SARNet~\cite{qin2024sarnet} employ multi-modal fusion or semantic constraints.
However, these approaches either require additional modalities such as RGB-D data or calibrated images, or rely on explicit semantic labels produced by task-specific segmentation models.
While these methods demonstrate the value of semantic information, directly integrating frozen pre-trained semantic representations into point cloud registration without additional sensors or retraining remains unexplored.
To address this, we propose an architecture that leverages frozen semantic representations solely from point clouds.

\vspace{3pt}\noindent\textbf{Mixture of experts.}
Mixture of Experts (MoE) offers an architectural approach to handle diverse input patterns efficiently.
Specifically, this approach has a set of small modules, called experts, and each expert specializes in different input patterns~\cite{shazeer2017outrageously,fedus2022switch,zha2025pointmode}.
PSReg~\cite{huang2025psreg} pioneered the application of MoE to registration.
The method leverages GeoTransformer~\cite{qin2022geotransformer} to determine which regions overlap between point clouds, and uses this overlap information for routing inputs to experts.
By enabling different experts to specialize in different geometric patterns (e.g., planar vs. curved surfaces, indoor vs. outdoor structures), MoE provides adaptive processing capability.
However, routing solely on geometric features fundamentally limits expert specialization; while experts can handle different geometric patterns, they cannot distinguish between semantically different structures that exhibit similar geometry.
Our method addresses this limitation by incorporating pre-trained semantic representations into MoE routing, enabling experts to specialize according to both geometric and semantic characteristics.

\section{Method}
\label{sec:method}

\subsection{Problem Formulation}
\label{subsec:overview}

Given two partially overlapping point clouds, $\mathcal{P} = \{\mathbf{p}_i\}_{i=1}^{N_{\mathcal{P}}}$ and $\mathcal{Q} = \{\mathbf{q}_j\}_{j=1}^{N_{\mathcal{Q}}}$, where $\mathbf{p}_i, \mathbf{q}_j \in \mathbb{R}^3$ represent 3D point positions, the goal is to estimate the rigid transformation $\mathbf{T} = (\mathbf{R}, \mathbf{t})$ that aligns $\mathcal{Q}$ to $\mathcal{P}$, where $\mathbf{R} \in SO(3)$ is the rotation in $\mathbb{R}^3$ and $\mathbf{t} \in \mathbb{R}^3$ is the translation.

As shown in Fig.~\ref{fig:pipeline}, our proposed pipeline is based on GeoTransformer~\cite{qin2022geotransformer} but incorporates a newly developed semantic representation extractor and an SG-MoE Transformer.
The pipeline proceeds in five stages: 
(1) semantic representation extraction using a frozen Sonata encoder in Sec.~\ref{subsec:feature_extraction}, 
(2) geometric feature extraction with semantic representations as input to KPConv~\cite{thomas2019kpconv} in Sec.~\ref{subsec:feature_extraction}, 
(3) spatial aggregation of semantic representations to the superpoint level in Sec.~\ref{subsec:aggregation}, 
(4) alternating geometric self-attention and semantic-guided MoE cross-attention for $N$ iterations in Sec.~\ref{subsec:sg_moe},
and (5) correspondence establishment and registration~\cite{qin2022geotransformer}. 

\subsection{Two-Stage Feature Extraction}
\label{subsec:feature_extraction}

To efficiently extract semantic-geometric features from the inputs $\mathcal{P}$ and $\mathcal{Q}$, we adopt a two-stage strategy.
In the first stage, we employ Sonata~\cite{wu2025sonata} to extract point-level semantic representations $\mathbf{F}_s^{\mathcal{P}} \in \mathbb{R}^{N_{\mathcal{P}} \times d_s}$ from the input point cloud $\mathcal{P}$, yielding $\mathbf{F}_s^{\mathcal{P}} = \text{Sonata}(\mathcal{P})$.
Similarly, we obtain $\mathbf{F}_s^{\mathcal{Q}} \in \mathbb{R}^{N_{\mathcal{Q}} \times d_s}$ from $\mathcal{Q}$ in the same manner.
In the second stage, we use a trainable KPConv encoder~\cite{thomas2019kpconv} to obtain downsampled points $\hat{\mathcal{P}} = \{ \hat{{\bf p}}_i \}_{i=1}^{\hat{N}_{\mathcal{P}}}$ ($\hat{N}_{\mathcal{P}} \ll N_{\mathcal{P}}$) and their corresponding semantic-geometric features $\hat{\mathbf{F}}_{sg}^{\mathcal{P}} \in \mathbb{R}^{\hat{N}_{\mathcal{P}} \times d_{sg}}$ as $\big(\hat{\mathcal{P}},\hat{\mathbf{F}}_{sg}^{\mathcal{P}} \big) = \text{KPConv}\big(\mathcal{P}, \mathbf{F}_s^{\mathcal{P}}\big)$.
Similarly, we get $\big(\hat{\mathcal{Q}},\hat{\mathbf{F}}_{sg}^{\mathcal{Q}} \big) = \text{KPConv}\big(\mathcal{Q}, \mathbf{F}_s^{\mathcal{Q}}\big)$.

This two-stage design achieves semantic-geometric feature fusion by leveraging KPConv's ability to learn conditional convolutions.
By providing semantic representations $\mathbf{F}_s^{\mathcal{P}}$ and $\mathbf{F}_s^{\mathcal{Q}}$ as initial features, we enable KPConv's kernel weights to adapt to the joint semantic-geometric space.
The resulting features $\hat{\mathbf{F}}_{sg}^{\mathcal{P}}$ and $\hat{\mathbf{F}}_{sg}^{\mathcal{Q}}$ encode semantic-aware geometric patterns that can distinguish geometrically similar structures based on their semantic meaning.

\subsection{Spatial Feature Aggregation}
\label{subsec:aggregation}

Following GeoTransformer~\cite{qin2022geotransformer}'s coarse-to-fine strategy, we perform transformer processing iteratively to refine the extracted features above $\mathbf{F}_s^{\mathcal{P}}$ and $\hat{\mathbf{F}}_{sg}^{\mathcal{P}}$.
To process these features with the transformer architecture, we aggregate $\mathbf{F}_s^{\mathcal{P}}$ with $N_{\mathcal{P}}$ points to match the $\hat{N}_{\mathcal{P}}$ points of $\hat{\mathbf{F}}_{sg}^{\mathcal{P}}$.

For each point $\hat{\mathbf{p}}_k \in \hat{\mathcal{P}}$, we obtain its downsampled semantic representation $\hat{\mathbf{f}}_{s,k}^{\mathcal{P}}$ by average-pooling the features of its $K$ nearest neighbors ($K=16$) in the original point cloud $\mathcal{P}$ (i.e., $\hat{\mathbf{f}}_{s,k}^{\mathcal{P}} = \frac{1}{K} \sum_{i \in \mathcal{I}_k} \mathbf{f}_{s,i}^{\mathcal{P}}$). This yields the downsampled representations $\hat{\mathbf{F}}_s^{\mathcal{P}}$ with $\hat{N}_{\mathcal{P}}$ points, matching the spatial dimension of $\hat{\mathbf{F}}_{sg}^{\mathcal{P}}$. The same aggregation is applied to $\mathcal{Q}$ to obtain $\hat{\mathbf{F}}_s^{\mathcal{Q}}$.

\subsection{Semantic-Guided Mixture of Experts}
\label{subsec:sg_moe}

After the above aggregation, we perform the transformer-based refinement process iteratively for the semantic-geometric features $\hat{\mathbf{F}}_{sg}^{\mathcal{P}}$ and $\hat{\mathbf{F}}_{sg}^{\mathcal{Q}}$. In this process, we use a new MoE module called Semantic-Guided Mixture of Experts (SG-MoE).
Following GeoTransformer~\cite{qin2022geotransformer}, we refine semantic-geometric features through $N$ iterations of transformer blocks, where each block consists of a Geometric Transformer layer (self-attention) and an SG-MoE Transformer layer (cross-attention), followed by superpoint matching for correspondence establishment, a coarse-to-fine strategy that has proven effective in recent transformer-based registration methods~\cite{yu2021cofinet,huang2021predator,yu2023peal}.
Geometric Transformer processes each point cloud independently with geometric positional encodings, while SG-MoE refines features through cross-attention with semantic-guided expert routing.

Although PSReg~\cite{huang2025psreg} pioneered MoE for registration, its geometry-only routing cannot distinguish regions with similar geometry but different semantics. 
Moreover, if we directly use semantic-geometric features $\hat{\mathbf{F}}_{sg}^\mathcal{P}$ for routing, registration losses that optimize for geometric alignment cause these representations to collapse into purely geometric patterns, losing their semantic discriminability.
To overcome this issue, we design our SG-MoE to route experts using superpoint-level semantic representations $\hat{\mathbf{F}}_{s}^\mathcal{P}$ only.

\begin{figure*}[t]
  \centering
  \begin{minipage}{0.45\textwidth}
    \centering
    \includegraphics[width=0.8\linewidth]{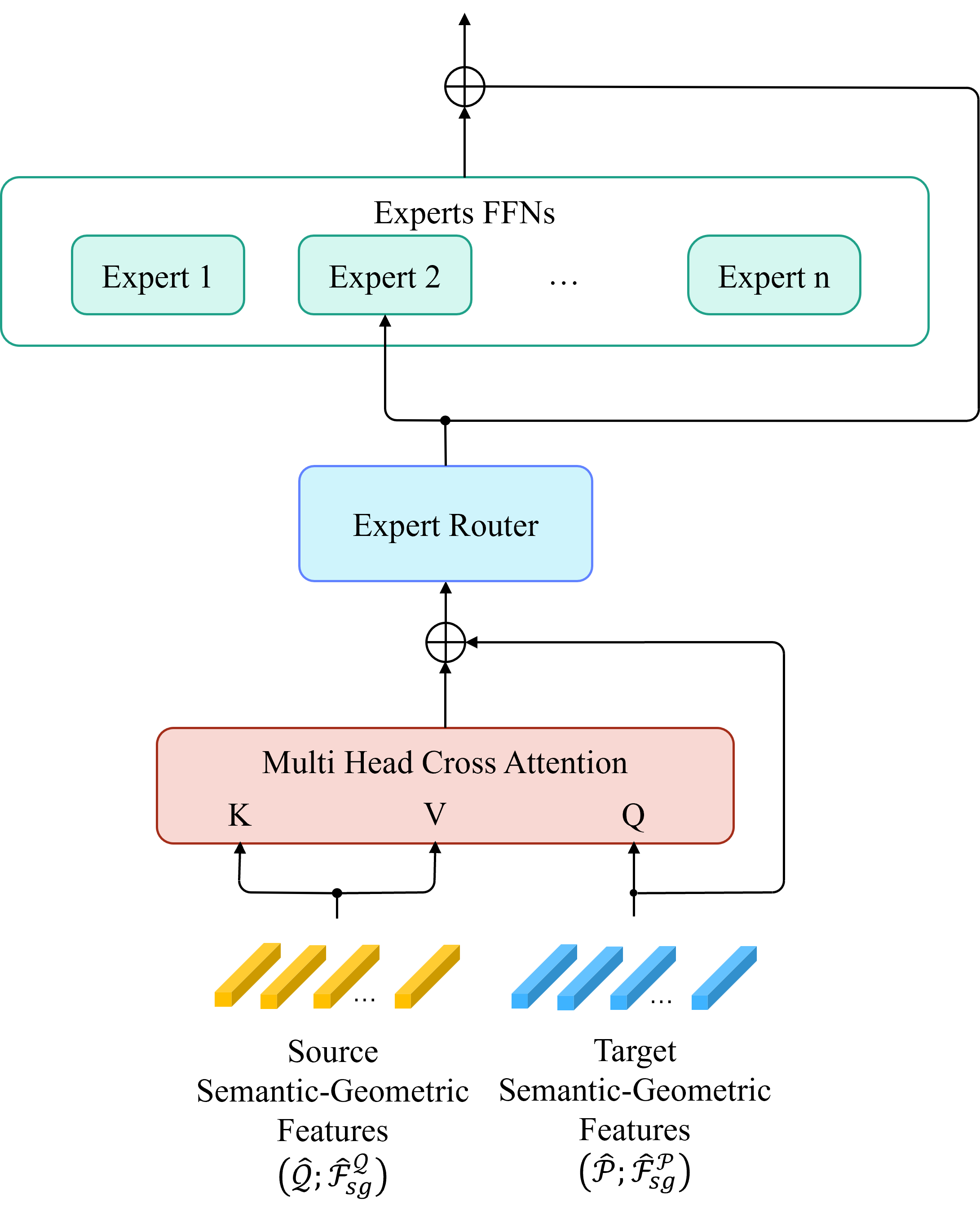}
    \subcaption{Standard MoE}
    \label{fig:left_image}
  \end{minipage}
  \hfill
  \begin{minipage}{0.45\textwidth}
    \centering
    \includegraphics[width=0.85\linewidth]{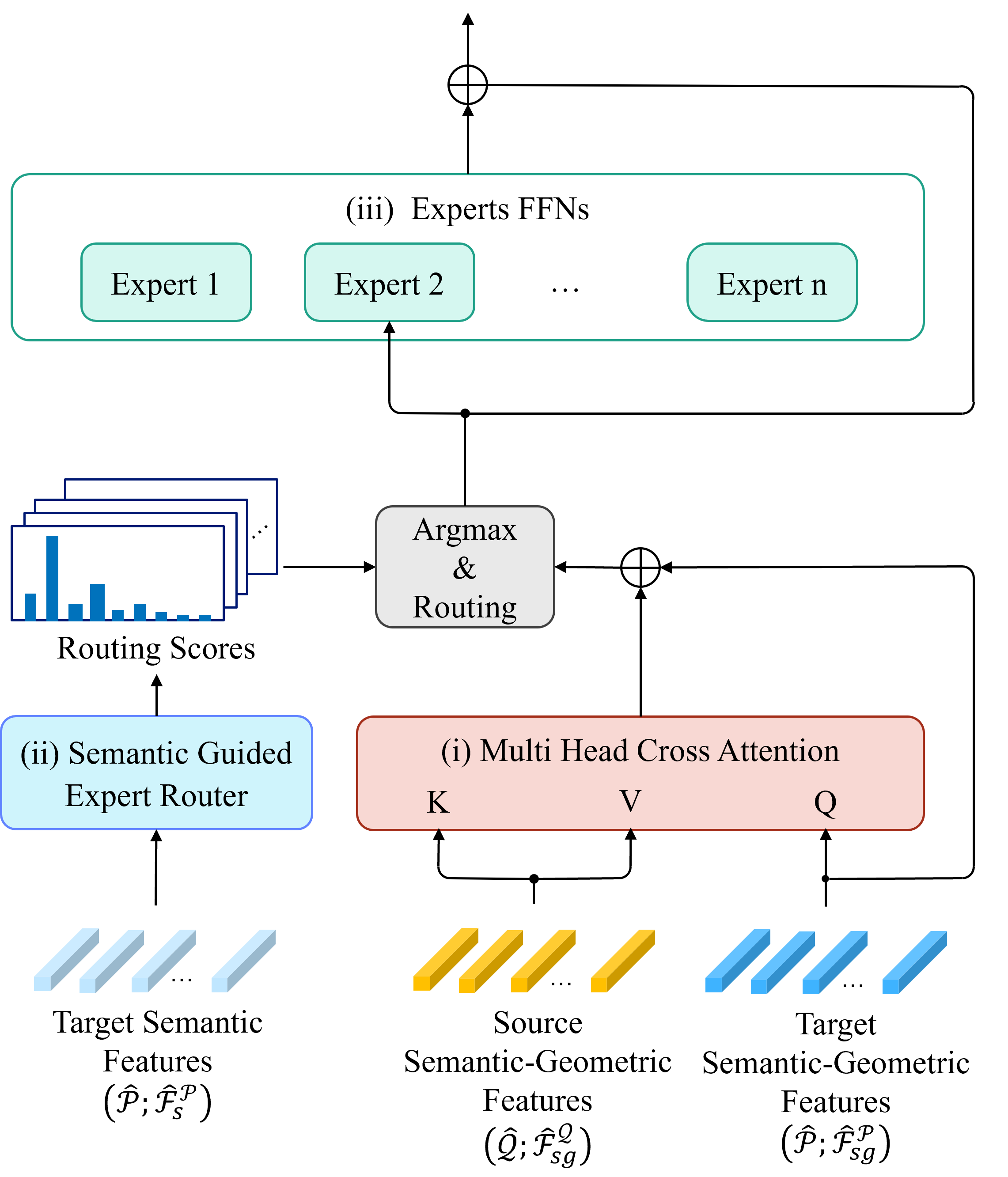}
    \subcaption{SG-MoE (ours)}
    \label{fig:right_image}
  \end{minipage}
  \caption{\textbf{Comparison of MoE architectures}. (a) Standard MoE feeds the fused semantic-geometric features into both the router and the experts. (b) Our SG-MoE decouples the inputs, feeding pure semantic features to the router and semantic-geometric features to the experts. These processes apply symmetrically with $\mathcal{P}$ and $\mathcal{Q}$ swapped.}
  \vspace{-12pt}
  \label{fig:sg_moe_layer}
  
\end{figure*}

As shown in Fig.~\ref{fig:sg_moe_layer}, SG-MoE consists of (i) a cross-attention layer, (ii) a semantic-guided router, and (iii) expert feed-forward networks (FFNs). 
Unlike Standard MoE (Fig.~\ref{fig:sg_moe_layer}a) whose fused-feature routing suffers from geometric domination, our SG-MoE (Fig.~\ref{fig:sg_moe_layer}b) strictly routes using pure semantic features. 
This design preserves semantic distinctions while experts refine the semantic-geometric features.

The refinement process in SG-MoE operates as follows.
First, a cross-attention layer updates semantic-geometric features:
{\small
\begin{equation}
    \hat{\mathbf{F}}^{\mathcal{P}}_{\text{attention}} =
    \text{LayerNorm}\left(\hat{\mathbf{F}}_{sg}^{\mathcal{P}} + \text{CrossAttn}(\hat{\mathbf{F}}_{sg}^{\mathcal{P}}, \hat{\mathbf{F}}_{sg}^{\mathcal{Q}})\right),
\end{equation}
}
Similarly, we get $\hat{\mathbf{F}}^{\mathcal{Q}}_{\text{attention}}$ from $\hat{\mathbf{F}}_{sg}^{\mathcal{Q}}$ and $\hat{\mathbf{F}}_{sg}^{\mathcal{P}}$.

In a semantic-guided expert router network, expert routing is guided by frozen semantic representations. 
For each superpoint's semantic representation $\hat{\mathbf{f}}_{s,k}^{\mathcal{P}} \in \mathbb{R}^{d_s}$, a lightweight two-layer routing network computes expert selection probabilities:
{\small
\begin{equation}
\mathbf{r}_k = \text{Softmax}(\pi(\hat{\mathbf{f}}_{s,k}^{\mathcal{P}})) \in \mathbb{R}^{N_{\text{expert}}}
\end{equation}
}
where $\pi$ is a two-layer MLP with ReLU activation.
Here, $\mathbf{r}_k = [r_{k,1}, \ldots, r_{k, N_{\text{expert}}}]$ represents the probability distribution over $N_{\text{expert}}$ experts for superpoint $k$, where $r_{k,i}$ indicates the probability of assigning superpoint $k$ to expert $i$. 
We employ a top-1 routing strategy where each superpoint is assigned to the expert with the highest probability, i.e., $i_k^* = \arg\max_{1 \le i \le N_{\text{expert}}} r_{k,i}$.

For each point $\hat{\mathbf{p}}_k$, let $\hat{\mathbf{f}}_{\text{attention},k}^{\mathcal{P}} \in \mathbb{R}^{d_{sg}}$ denote the $k$-th transposed row of $\hat{\mathbf{F}}^{\mathcal{P}}_{\text{attention}}$.
The semantic-geometric features from cross-attention are processed by the assigned expert:
{\small
\begin{equation}
    \hat{\mathbf{f}}_{\text{out},k}^{\mathcal{P}} = \text{LayerNorm}\left(\hat{\mathbf{f}}_{\text{attention},k}^{\mathcal{P}} + \psi_{i_k^*}(\hat{\mathbf{f}}_{\text{attention},k}^{\mathcal{P}})\right),
\end{equation}
}
where $\psi_{i_k^*}$ is a two-layer feed-forward network with ReLU activation.

The router is regularized by the load-balancing objective, while the selected experts are optimized by the registration losses.
Because routing is determined from frozen semantic features, different experts receive different subsets of semantic-geometric features and can consequently learn distinct refinement functions.
This semantic-conditioned expert specialization helps reduce ambiguities that have been reported as a limitation of geometry-only registration methods~\cite{qin2022geotransformer, yu2023roitr}.

Following GeoTransformer~\cite{qin2022geotransformer}, after $N$ transformer block iterations, the refined features are used to establish correspondences through superpoint matching and point matching, followed by transformation estimation.

\subsection{Training Objective}
\label{subsec:training}

We train SAGE end-to-end using:
\begin{equation}
\mathcal{L}_{\text{total}} = \lambda_{\text{coarse}} \mathcal{L}_{\text{coarse}} + \lambda_{\text{fine}} \mathcal{L}_{\text{fine}} + \lambda_{\text{moe}} \mathcal{L}_{\text{moe}}.
\label{eq:loss}
\end{equation}
The first two terms supervise correspondence quality at different spatial resolutions, while the third term regularizes expert utilization in the SG-MoE layer. We describe each term below.
Note that the Sonata encoder remains frozen throughout training.
Its semantic representations are never updated by gradients, preserving pre-trained knowledge. 
Therefore, the KPConv encoder, transformer layers, routing networks, and matching heads are trained by the loss~\eqref{eq:loss}.

\vspace{3pt}\noindent\textbf{Coarse-level supervision loss} $(\mathcal{L}_{\text{coarse}})$.
We adopt the overlap-aware circle loss from GeoTransformer~\cite{qin2022geotransformer} to supervise superpoint correspondences.
The loss is computed bidirectionally as $\mathcal{L}_{\text{coarse}} = \mathcal{L}_{\text{oc}}^{\mathcal{P} \to \mathcal{Q}} + \mathcal{L}_{\text{oc}}^{\mathcal{Q} \to \mathcal{P}}$, where $\mathcal{L}_{\text{oc}}$ denotes this loss computed in each direction\footnote{The detailed definition is introduced in Appendix B.1.}.

\vspace{3pt}\noindent\textbf{Fine-level supervision loss} $(\mathcal{L}_{\text{fine}})$.
Following the optimal transport formulation in GeoTransformer~\cite{qin2022geotransformer}, we supervise point-level matches within each superpoint pair using Sinkhorn iterations.
Here, $\tilde{\mathcal{P}}$ and $\tilde{\mathcal{Q}}$ denote intermediate-resolution downsampled points obtained from the KPConv encoder.
During training, we randomly sample $N_g=128$ ground-truth superpoint pairs.
For each pair, we compute matching costs from point features at the $\tilde{\mathcal{P}}$, $\tilde{\mathcal{Q}}$ resolution and obtain soft assignments $\mathbf{A}_i$ through the Sinkhorn algorithm, whose $(p,q)$-th entry $a_{p,q}^i$ represents the matching probability between two points.
The negative log-likelihood loss is:
\begin{equation}
\mathcal{L}_{\text{fine}} = \frac{1}{N_g} \sum_{i=1}^{N_g} \left( -\sum_{(p,q) \in \Gamma_i} \log a_{p,q}^i \right)
\end{equation}
where $\Gamma_i$ includes ground-truth matches (within radius 0.05\,m) and dustbin assignments for unmatched points\footnote{The detailed definition is introduced in Appendix B.2.}.

\vspace{3pt}\noindent\textbf{Expert load balancing loss} $(\mathcal{L}_{\text{moe}})$.
To prevent the routing network from collapsing to a subset of experts, we adopt the load balancing loss~\cite{shazeer2017outrageously}. Specifically, we compute the average routing probability $\bar{r}_e = \frac{1}{|\mathcal{B}|}\sum_{k \in \mathcal{B}} r_{k,e}$ for each expert $e$ across all superpoints in the batch $\mathcal{B}$ and minimize its KL divergence from the uniform distribution:
{
\begin{equation}
\mathcal{L}_{\text{moe}} = \sum_{e=1}^{N_{\text{expert}}} 
\bar{r}_e \log \frac{\bar{r}_e}{1/N_{\text{expert}}}.
\end{equation}
}
\section{Experiments}
\label{sec:experiments}

\subsection{Implementation Details}
\label{subsec:impl}

\vspace{3pt}\noindent\textbf{Network architecture.}
We use $N=3$ transformer block iterations. 
The number of experts is set to $N_{\text{expert}}=4$. For the pair-wise distance embedding in geometric self-attention, we set $\sigma_d$ to the voxel size at the superpoint level, and $\sigma_a = 15^\circ$ for the triplet-wise angular embedding.
For local-to-global registration, only superpoint matches with at least 3 local point correspondences are used to compute transformation candidates. 
We iteratively recompute the transformation with the surviving inlier matches $N_r=5$ times.

\vspace{3pt}\noindent\textbf{Training and testing.}
We implement and train SAGE with PyTorch on an NVIDIA H100 GPU. Models are trained with the Adam optimizer for 45 epochs on 3DMatch with a batch size of 1. 
The learning rate starts from $10^{-4}$ and is multiplied by 0.95 after each epoch.
Loss weights are $\lambda_{\text{coarse}} = 1.0$, $\lambda_{\text{fine}} = 1.0$ and $\lambda_{\text{moe}} = 0.01$. We randomly sample $N_g=128$ ground-truth superpoint correspondences during training, and $N_c=256$ putative superpoint matches during testing. Data augmentation follows~\cite{huang2021predator}.

\vspace{3pt}\noindent\textbf{Baselines and reproduction.}
In the experiments, we evaluate two approaches: a single-pass approach and an iterative refinement approach.
The former predicts the correspondences and the transformation directly from the extracted features without any feedback, and the latter iteratively updates the correspondences by embedding the estimate from the previous iteration as a prior to guide the subsequent matching.
In the single-pass approach, we implemented several well-known methods, Predator~\cite{huang2021predator}, CoFiNet~\cite{yu2021cofinet}, GeoTransformer~\cite{qin2022geotransformer} and RoITr~\cite{yu2023roitr}, to compare with our SAGE. 
For the iterative refinement approach, we extended our SAGE with the same iterative refinement used in PEAL~\cite{yu2023peal} and PSReg~\cite{huang2025psreg}.
We call this extension SAGE$^\dagger$ in this paper based on the official implementation of PEAL~\cite{yu2023peal_code}.
On the other hand, the implementation of PSReg is not officially released, and several implementation hyperparameters are not reported, so we re-implemented PSReg from scratch.
We restrict the comparison to methods that operate on point clouds alone, and therefore exclude semantic-enhanced methods such as SemReg~\cite{fung2024semreg}, which require additional RGB images from calibrated cameras.

\subsection{Evaluation on 3DMatch and 3DLoMatch}
We evaluate SAGE on the 3DMatch~\cite{zeng20173dmatch} and 3DLoMatch~\cite{huang2021predator} datasets to assess its performance in point cloud registration.
We evaluate performance using three metrics following~\cite{huang2021predator,qin2022geotransformer}: \textit{Inlier Ratio (IR)}, which measures the fraction of putative correspondences whose residuals are below 0.1\,m under the ground-truth transformation; \textit{Feature Matching Recall (FMR)}, which calculates the percentage of point cloud pairs with IR exceeding 5\%; and \textit{Registration Recall (RR)}, which determines the fraction of pairs whose transformation error (RMSE) is below 0.2\,m\footnote{The definition of evaluation metrics is introduced in Appendix C.1.}.

\vspace{3pt}\noindent\textbf{Results with RANSAC-50K.}
%We compare recent methods.
Following standard protocols~\cite{huang2021comprehensive}, we compare recent point cloud registration methods with our SAGE. 
We evaluate each method using RANSAC-50K with variable sample counts. Table~\ref{tab:main} presents comprehensive results on both 3DMatch and 3DLoMatch benchmarks. 

\definecolor{multipasspink}{RGB}{255, 220, 230}
\begin{table}
\centering
\setlength{\tabcolsep}{4.2pt}
\resizebox{\linewidth}{!}{
\begin{tabular}{l|ccccc|ccccc}
\toprule
& \multicolumn{5}{c|}{\textbf{3DMatch}} & \multicolumn{5}{c}{\textbf{3DLoMatch}} \\
\# Samples & 5000 & 2500 & 1000 & 500 & 250 & 5000 & 2500 & 1000 & 500 & 250 \\
\midrule
\multicolumn{11}{c}{\textit{Feature Matching Recall (\%) $\uparrow$}} \\
\midrule
Predator~\cite{huang2021predator} & 96.6 & 96.6 & 96.5 & 96.3 & 96.5 & 78.6 & 77.4 & 76.3 & 75.7 & 75.3 \\
CoFiNet~\cite{yu2021cofinet} & \underline{98.1} & \underline{98.3} & \underline{98.1} & \underline{98.2} & \textbf{98.3} & 83.1 & 83.5 & 83.3 & 83.1 & 82.6 \\
GeoTrans.~\cite{qin2022geotransformer} & 97.9 & 97.9 & 97.9 & 97.9 & 97.6 & \underline{88.3} & 88.6 & 88.8 & 88.6 & 88.3 \\
RoITr~\cite{yu2023roitr} & 98.0 & 98.0 & 97.9 & 98.0 & \underline{97.9} & \textbf{89.6} & \underline{89.6} & \underline{89.5} & \underline{89.4} & \underline{89.3} \\
SAGE (\textit{ours}) & \textbf{98.4} & \textbf{98.5} & \textbf{98.6} & \textbf{98.6} & \textbf{98.3} & \textbf{89.6} & \textbf{89.7} & \textbf{89.9} & \textbf{90.0} & \textbf{89.4} \\
\rowcolor{multipasspink} PEAL~\cite{yu2023peal} & \textbf{97.8} & \textbf{98.1} & \textbf{98.0} & \textbf{98.0} & \textbf{97.8} & \textbf{85.8} & \textbf{86.0} & \underline{85.5}& 85.1 & 85.0 \\
\rowcolor{multipasspink} PSReg~\cite{huang2025psreg} & 97.1 & 97.4 & \underline{97.7}& 97.7 & \underline{97.7}& 85.3 & 85.7 & \textbf{85.7}& \textbf{85.8}& \textbf{85.3} \\
\rowcolor{multipasspink} SAGE$^\dagger$ (\textit{ours}) & \underline{97.2}& \underline{97.5}& \underline{97.7} & \underline{97.8}& \underline{97.7}& \underline{85.4}& \underline{85.9}& 85.4 & \underline{85.2}& \textbf{85.3} \\
\midrule
\multicolumn{11}{c}{\textit{Inlier Ratio (\%) $\uparrow$}} \\
\midrule
Predator~\cite{huang2021predator} & 58.0 & 58.4 & 57.1 & 54.1 & 49.3 & 26.7 & 28.1 & 28.3 & 27.5 & 25.8 \\
CoFiNet~\cite{yu2021cofinet} & 49.8 & 51.2 & 51.9 & 52.2 & 52.2 & 24.4 & 25.9 & 26.7 & 26.8 & 26.9 \\
GeoTrans.~\cite{qin2022geotransformer} & 71.9 & 75.2 & 76.0 & 82.2 & \underline{85.1} & 43.5 & 45.3 & 46.2 & 52.9 & \underline{57.7} \\
RoITr~\cite{yu2023roitr} & \textbf{82.6} & \textbf{82.8} & \underline{83.0} & \underline{83.0} & 83.0 & \textbf{54.3} & \textbf{54.6} & \underline{55.1} & \underline{55.2} & 55.3 \\
SAGE (\textit{ours}) & \underline{73.5} & \underline{79.8} & \textbf{84.1} & \textbf{85.8} & \textbf{87.0} & \underline{44.5} & \underline{50.3} & \textbf{56.4} & \textbf{59.1} & \textbf{60.9} \\
\rowcolor{multipasspink} PEAL~\cite{yu2023peal} & 74.9 & \underline{81.3} & \underline{85.7} & \textbf{87.5} & \textbf{88.7} & \underline{49.7} & \underline{55.0} & \underline{61.1} & \textbf{63.6} & \textbf{65.2} \\
\rowcolor{multipasspink} PSReg~\cite{huang2025psreg} & \underline{75.0} & \underline{81.3} & \underline{85.7} & \underline{87.4} & 88.5 & 49.5 & \underline{55.0} & 61.0 & \underline{63.5} & \underline{65.1} \\
\rowcolor{multipasspink} SAGE$^\dagger$ (\textit{ours}) & \textbf{75.5} & \textbf{81.7} & \textbf{85.9} & \textbf{87.5} & \underline{88.6} & \textbf{49.9} & \textbf{55.4} & \textbf{61.2} & \textbf{63.6} & \underline{65.1} \\
\midrule
\multicolumn{11}{c}{\textit{Registration Recall (\%) $\uparrow$}} \\
\midrule
Predator~\cite{huang2021predator} & 89.0 & 89.9 & 90.6 & 88.5 & 86.6 & 59.8 & 61.2 & 62.4 & 60.8 & 58.1 \\
CoFiNet~\cite{yu2021cofinet} & 89.3 & 88.9 & 88.4 & 87.4 & 87.0 & 67.5 & 66.2 & 64.2 & 63.1 & 61.0 \\
GeoTrans.~\cite{qin2022geotransformer} & \textbf{92.0} & \underline{91.8} & \underline{91.8} & \textbf{91.4} & \textbf{91.2} & \textbf{75.0} & \textbf{74.8} & 74.2 & 74.1 & 73.5 \\
RoITr~\cite{yu2023roitr} & \underline{91.9} & 91.7 & \underline{91.8} & \textbf{91.4} & \underline{91.0} & \underline{74.7} & \textbf{74.8} & \textbf{74.8} & \underline{74.2} & \underline{73.6} \\
SAGE (\textit{ours}) & 90.8 & \textbf{92.0} & \textbf{91.9} & \underline{91.1} & \underline{91.0} & 73.4 & \underline{74.2} & \underline{74.5} & \textbf{74.6} & \textbf{73.9} \\
\rowcolor{multipasspink} PEAL~\cite{yu2023peal} & \underline{91.9} & 91.6 & 91.7 & 91.7 & 91.7 & 76.1 & 76.1 & 76.2 & \underline{75.9} & \underline{75.9} \\
\rowcolor{multipasspink} PSReg~\cite{huang2025psreg} & \underline{91.9} & \underline{91.9} & \underline{92.2} & \textbf{92.0} & \underline{91.8}  & \underline{76.4} & \underline{76.5} & \underline{76.4} & \underline{75.9} & 75.3 \\
\rowcolor{multipasspink} SAGE$^\dagger$ (\textit{ours}) & \textbf{92.2} & \textbf{92.3} & \textbf{92.5} & \underline{91.9} & \textbf{92.1} & \textbf{77.3} & \textbf{77.3} & \textbf{77.0} & \textbf{76.3} & \textbf{76.0} \\
\bottomrule
\end{tabular}
}
\caption{Evaluation results on 3DMatch and 3DLoMatch using RANSAC-50K. \colorbox{multipasspink}{Pink rows} use iterative refinement. \textbf{Boldface} indicates the best result and \underline{underline} indicates the second-best result within each group. Iterative PEAL and PSReg are reproduced under our setting and fall below their reported numbers.}
\vspace{-12pt}
\label{tab:main}
\end{table}

As shown in Table~\ref{tab:main}, among the single-pass methods the best and second-best scores are highlighted in boldface and underline, respectively.
SAGE achieves the best FMR across all sample counts (98.3--98.6\%), demonstrating a superior feature matching capability in high-overlap scenarios. 
For IR, SAGE achieves the best performance with 1000 or fewer samples, reaching 87.0\% at 250 samples, and is second only to RoITr at 5000 and 2500 samples. 
For RR, SAGE achieves 92.0\% at 2500 samples and 91.9\% at 1000 samples, both of which are the highest among single-pass methods, while remaining competitive at other sample counts.
The slight gap at 5000 samples (90.8\% vs. GeoTransformer's 92.0\%) is due to a limitation of RANSAC discussed below.

On 3DLoMatch, SAGE achieves the best or tied-best FMR across all sample counts (89.6--90.0\%). For IR, SAGE achieves the best performance with 1000 or fewer samples (56.4--60.9\%), outperforming RoITr's 55.3\% by 5.6 points at 250 samples, and is second-best at 5000 and 2500 samples. 
This demonstrates SAGE's robustness with limited correspondences. 
For RR, SAGE achieves 73.4--74.6\%, which is competitive but slightly below the best methods in some configurations, due to the same limitation of RANSAC.

Specifically, the limitation of RANSAC in our setting is that its uniform random sampling treats all correspondences equally, failing to exploit the semantic structure that SAGE introduces.

\vspace{3pt}\noindent\textbf{Results with structure-aware estimators.}
We next evaluate SAGE under two protocols that overcome the limitation of RANSAC, and we use these methods to directly compare SAGE with the most relevant recent methods.
The upper group of Table~\ref{tab:rr} reports single-pass results with LGR~\cite{qin2022geotransformer}, a structure-aware estimator that exploits spatial relationships among correspondences rather than sampling them uniformly.
Because PEAL refines its estimate over several iterations, we evaluate SAGE, PEAL, and PSReg under a common iterative protocol in which each method receives the same budget of six refinement iterations, which places them under identical input and evaluation conditions.

\begin{table}
\centering
\setlength{\tabcolsep}{4.2pt}
\renewcommand{\arraystretch}{1.1}
\resizebox{0.75\linewidth}{!}{
\begin{tabular}{l|ccc|ccc}
\toprule
& \multicolumn{3}{c|}{\textbf{3DMatch}} & \multicolumn{3}{c}{\textbf{3DLoMatch}} \\
& FMR$\uparrow$ & IR$\uparrow$ & RR$\uparrow$ & FMR$\uparrow$ & IR$\uparrow$ & RR$\uparrow$ \\
\midrule
GeoTrans.~\cite{qin2022geotransformer} & \textbf{98.2} & 70.9 & \textbf{92.5} & \underline{87.1}& 43.5 & \underline{74.0}\\
RoITr~\cite{yu2023roitr} & \underline{98.1}& \textbf{82.0} & 90.9 & \textbf{89.4} & \textbf{54.4} & 73.4 \\
SAGE (\textit{ours}) & \textbf{98.2} & \underline{71.7}& \underline{91.0}& \textbf{89.4} & \underline{44.3}& \textbf{76.3} \\
\rowcolor{multipasspink} PEAL~\cite{yu2023peal} & \textbf{97.8} & \underline{73.5}& 91.5 & \textbf{85.8} & \underline{49.5}& 76.9 \\
\rowcolor{multipasspink} PSReg~\cite{huang2025psreg} & 97.1 & 73.4 & \underline{92.1}& \underline{85.4}& \underline{49.5}& \underline{77.4}\\
\rowcolor{multipasspink} SAGE$^\dagger$ (\textit{ours}) & \underline{97.2}& \textbf{73.7} & \textbf{92.2} & \underline{85.4}& \textbf{49.7} & \textbf{77.6} \\
\bottomrule
\end{tabular}
}
\caption{Evaluation on 3DMatch and 3DLoMatch with LGR~\cite{qin2022geotransformer}.} 
\vspace{-6pt}
\label{tab:rr}
\end{table}

Under LGR, SAGE attains the highest RR on 3DLoMatch (76.3\%), surpassing GeoTransformer by 2.3 points and RoITr by 2.9 points, while remaining competitive on 3DMatch (91.0\%).
SAGE reaches this highest RR on 3DLoMatch despite a lower IR than RoITr, which indicates that its correspondences exhibit stronger spatial consistency that structure-aware estimators can effectively exploit.

Under this protocol, SAGE attains the highest IR and the highest RR on both 3DMatch and 3DLoMatch, reaching 92.2\% and 77.6\% RR respectively, whereas PEAL records the highest FMR.
This shows that, when every method receives the same refinement budget, the semantic-guided routing in SAGE produces correspondences that are at least as reliable as those of the strongest recent MoE-based and iterative methods.

\subsection{Ablation Studies}

We analyze each component's contribution on 3DLoMatch using LGR. 
Results are shown in Table~\ref{tab:ablation}.

\begin{table}
\centering
\setlength{\tabcolsep}{4.2pt}
\resizebox{\linewidth}{!}{
\begin{tabular}{lccc}
\toprule
\textbf{Configuration} & \textbf{FMR} $\uparrow$ [\%] & \textbf{IR}$\uparrow$ [\%] & \textbf{RR}$\uparrow$ [\%] \\
\midrule
baseline & 87.1 & 43.5 & 74.0 \\
Two-Stage Feature Extraction only & \underline{88.5} & 40.4 & 74.7 \\
Standard MoE only & 87.5 & 42.6 & 74.7 \\
SG-MoE only  & 87.5 & \underline{43.8} & \underline{75.3} \\
\midrule
Full SAGE & \textbf{89.4} & \textbf{44.3} & \textbf{76.3} \\
\bottomrule
\end{tabular}
}
\caption{Ablation study on 3DLoMatch with LGR.}
\vspace{-12pt}
\label{tab:ablation}
\end{table}

\begin{figure*}[t]
\centering
\includegraphics[width=0.95\linewidth]{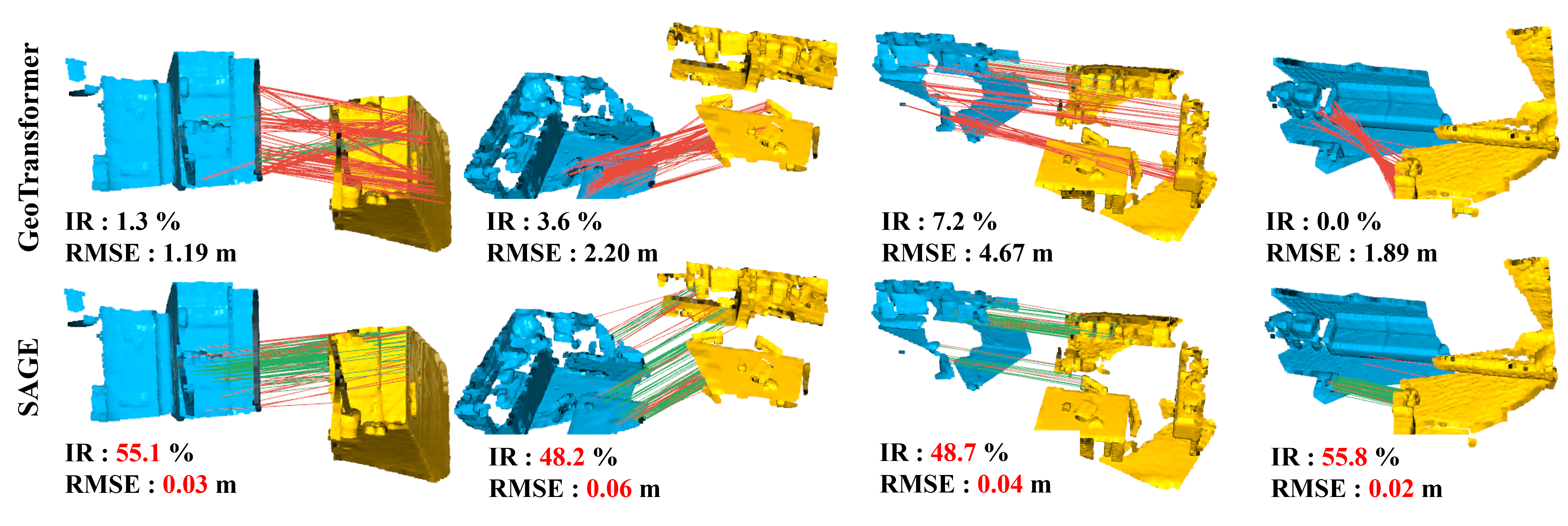}
\caption{\textbf{Qualitative comparison of correspondences on 3DLoMatch scenes. }
Each column shows a different point cloud pair. 
Top: GeoTransformer's registration with correspondences (red for outliers, green for inliers). 
Bottom: SAGE's registration and correspondences.
SAGE effectively leverages semantic guidance to correctly identify inliers, even in low-overlap scenarios. }
\vspace{-12pt}
\label{fig:qualitative}
\end{figure*}

\vspace{3pt}\noindent\textbf{Effect of SG-MoE.}
Removing SG-MoE decreases RR by 1.6 points and IR by 3.9 points. 
This demonstrates that semantic-guided expert routing directly improves correspondence quality. 
By routing based on frozen semantic features, experts naturally specialize for different semantic-geometric patterns, enabling effective disambiguation of geometrically similar but semantically distinct regions (e.g., walls vs. floors).

\vspace{3pt}\noindent\textbf{Effect of two-stage feature extraction.}
Without two-stage feature extraction, FMR drops 1.9 points and RR drops 1.0 point, while IR remains relatively stable. 
This indicates that semantic input primarily enables more discriminative geometric features by grounding them in semantic understanding, leading to better feature matching across more point cloud pairs.

\vspace{3pt}\noindent\textbf{Complementary effects.}
Comparing w/ two-stage feature extraction (RR: 74.7\%, IR: 40.4\%) and w/ SG-MoE (RR: 75.3\%, IR: 43.8\%) reveals that both components contribute to different aspects. 
Two-stage feature extraction mainly improves FMR (+1.4 points) at the cost of IR ($-$3.1 points), whereas SG-MoE improves both IR and RR (+0.3 and +1.3 points). 
The complete SAGE achieves a 2.3-point RR improvement over the baseline, demonstrating the synergy of both components. 
Even with only one component, SAGE outperforms the baseline in RR, confirming the value of each design choice.

\subsection{Qualitative Results}
Fig.~\ref{fig:qualitative} shows qualitative comparisons on challenging 3DLoMatch scenes with geometrically similar structures. 
We visualize correspondences as lines colored green for inliers and red for outliers. 
In scenes where some furniture shares similar local geometry, GeoTransformer~\cite{qin2022geotransformer} (top row) produces predominantly outliers, resulting in low IR in some cases. 
In contrast, SAGE (bottom row) successfully distinguishes these geometrically similar but semantically distinct structures through semantic understanding, achieving high IR and accurate final registration. 
This visual evidence shows that semantic-guided expert routing effectively handles ambiguous geometric patterns that confound purely geometric methods. 
\footnote{Additional qualitative comparisons with GeoTrans. are provided in Appendix D.6 (Fig.~B).}

\begin{figure}[t]
\centering
\includegraphics[width=0.85\linewidth]{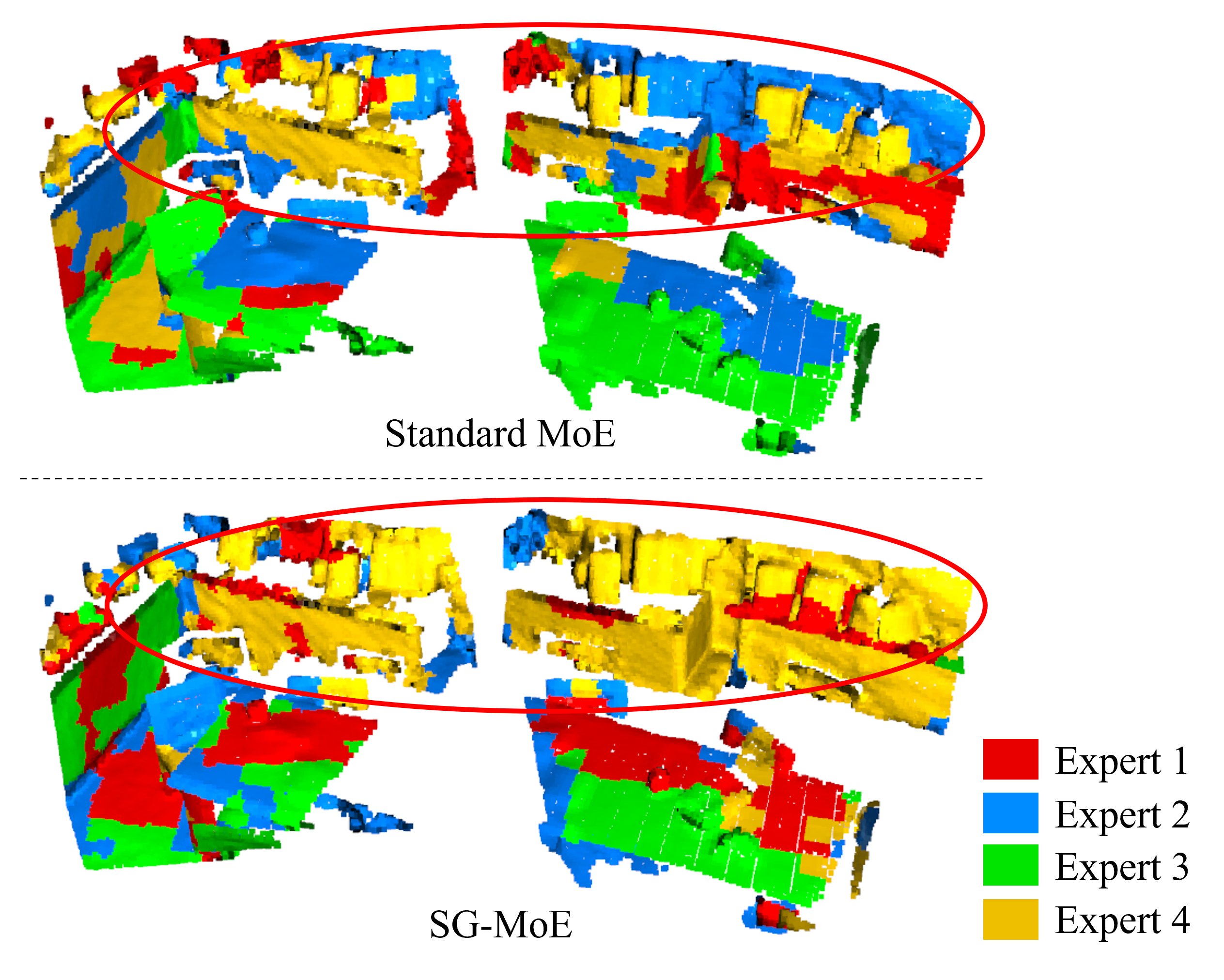}
\caption{\textbf{Expert activation patterns.}
\textbf{Top:} Standard MoE produces fragmented expert assignments primarily driven by local geometry.
\textbf{Bottom:} SG-MoE consistently assigns semantically similar regions, such as the furniture highlighted by red circles, to the same expert, demonstrating semantic-guided specialization.}
\vspace{-6pt}
\label{fig:routing_visualization}
\end{figure}

Fig.~\ref{fig:routing_visualization} compares the expert activation patterns of Standard MoE and SG-MoE.
Standard MoE produces fragmented assignments primarily driven by local geometric patterns, whereas SG-MoE assigns semantically similar regions to consistent experts, as illustrated by the coherent assignments of the highlighted furniture.
This specialization emerges without explicit supervision of expert roles and supports consistent feature processing across semantically related structures.
\section{Conclusion}

We presented SAGE, a framework that integrates frozen pre-trained semantic representations with a mixture of experts for point cloud registration, requiring neither additional sensors nor retraining of the semantic encoder.
SAGE addresses a key limitation of geometry-only methods by distinguishing geometrically similar but semantically different structures, and achieves the highest registration recall among directly comparable methods on low-overlap 3DLoMatch while remaining competitive on 3DMatch and outdoor scenes.
{
    \small
    \bibliographystyle{ieeenat_fullname}
    \bibliography{main}
}
\newpage
\maketitlesupplementary
\renewcommand{\thefigure}{\Alph{figure}}
\renewcommand{\thetable}{\Alph{table}}
\setcounter{figure}{0}
\setcounter{table}{0}

\appendix

\section{Network Architecture Details}
\label{sec:app_net}

\subsection{Semantic Representation Extractor (Sonata)}
\label{sec:a_sonata}
SAGE employs Sonata~\cite{wu2025sonata} as the semantic representation extractor.
Sonata is a point cloud foundation model that is pre-trained through a contrastive learning objective designed to align point cloud representations with semantically meaningful structures while explicitly avoiding geometric shortcuts.
Its encoder follows a sparse convolutional architecture that processes voxelized point clouds and produces per-point feature vectors of 1024 dimensions.

An important property of Sonata for our framework is that it outputs continuous, high-dimensional semantic representation vectors rather than discrete categorical labels.
Unlike conventional semantic segmentation models that predict specific categories such as ``chair'' or ``car,'' Sonata captures rich semantic relationships in a continuous feature space where semantically similar structures are grouped together without being constrained by a predefined label vocabulary.
This characteristic is essential for SAGE because it allows the SG-MoE router to exploit fine-grained semantic distinctions that go beyond coarse category boundaries.
For both indoor and outdoor environments, we use the frozen Sonata model exclusively as a feature extractor, which ensures domain-agnostic operation that remains entirely free from the constraints of dataset-specific label spaces.

\subsection{Geometric Feature Extractor (KPConv) for 3DMatch and 3DLoMatch}
\label{sec:a_1}
Following GeoTransformer~\cite{qin2022geotransformer}, we use a KPConv-FPN backbone for feature extraction.
Prior to feeding the data into the backbone network, the input point clouds are processed with an initial voxelization using a sampling resolution of 2.5\,cm. 
As the data progresses through successive downsampling stages, the voxel resolution is progressively coarsened by a factor of two at each level.

As shown in Table~\ref{tab:architecture}, SAGE adopts the same KPConv-FPN backbone as GeoTransformer~\cite{qin2022geotransformer}, with the only modification being the input dimension of the first KPConv layer.
Specifically, we expand the input feature dimension from 1 (no input features) to 1088 to incorporate pre-trained semantic representations from Sonata~\cite{wu2025sonata}.
This enables the backbone to learn semantic-geometric features from the initial stage.

\begin{table}

\caption{Geometric Feature Extractor (KPConv) architecture. SAGE modifies only the input dimension of the first KPConv layer (red) to accommodate 1024-dimensional Sonata semantic features concatenated with 64-dimensional positional encodings.}
\label{tab:architecture}
\centering
\setlength{\tabcolsep}{4.2pt}
\resizebox{\linewidth}{!}{
\begin{tabular}{c|c|c}
\toprule
Stage & \multicolumn{1}{c|}{GeoTransformer} & \multicolumn{1}{c}{SAGE} \\
\midrule
1 & KPConv($1 \rightarrow 64$) & KPConv(\textcolor{red}{1088} $\rightarrow 64$) \\
  & ResBlock($64 \rightarrow 128$) & ResBlock($64 \rightarrow 128$) \\
\midrule
2 & ResBlock($128 \rightarrow 128$, strided) & ResBlock($128 \rightarrow 128$, strided) \\
  & ResBlock($128 \rightarrow 256$) & ResBlock($128 \rightarrow 256$) \\
  & ResBlock($256 \rightarrow 256$) & ResBlock($256 \rightarrow 256$) \\
\midrule
3 & ResBlock($256 \rightarrow 256$, strided) & ResBlock($256 \rightarrow 256$, strided) \\
  & ResBlock($256 \rightarrow 512$) & ResBlock($256 \rightarrow 512$) \\
  & ResBlock($512 \rightarrow 512$) & ResBlock($512 \rightarrow 512$) \\
\midrule
4 & ResBlock($512 \rightarrow 512$, strided) & ResBlock($512 \rightarrow 512$, strided) \\
  & ResBlock($512 \rightarrow 1024$) & ResBlock($512 \rightarrow 1024$) \\
  & ResBlock($1024 \rightarrow 1024$) & ResBlock($1024 \rightarrow 1024$) \\
\midrule
\multicolumn{3}{c}{\textit{Upsampling (FPN)}} \\
\midrule
5 & NearestUpsampling & NearestUpsampling \\
  & UnaryConv($1536 \rightarrow 512$) & UnaryConv($1536 \rightarrow 512$) \\
\midrule
6 & NearestUpsampling & NearestUpsampling \\
  & UnaryConv($768 \rightarrow 256$) & UnaryConv($768 \rightarrow 256$) \\

\bottomrule
\end{tabular}
}

\end{table}

\subsection{Geometric Feature Extractor (KPConv) for KITTI}
\label{sec:a_2}
For the KITTI odometry benchmark, we adapt the backbone architecture to handle larger-scale outdoor environments.
The input point clouds are voxelized with a sampling resolution of 30\,cm, significantly coarser than the 2.5\,cm used for 3DMatch, reflecting the difference in spatial scale between indoor rooms and outdoor driving scenarios.

As shown in Table~\ref{tab:architecture_kitti}, we employ a deeper 5-stage backbone compared to the 4-stage backbone used for 3DMatch.
The additional downsampling stage (Stage 5) is necessary to capture long-range spatial relationships in outdoor scenes that can span over 100 meters, while maintaining computational efficiency.
Without this additional stage, the receptive field would be insufficient to establish correspondences across the large spatial extents typical of KITTI scenes.

Similar to 3DMatch, we modify only the input feature dimension of the first KPConv layer from 1 to 1088 to incorporate semantic features from Sonata~\cite{wu2025sonata}.
The deeper architecture with Feature Pyramid Network~\cite{lin2017fpn} connections enables effective multi-scale feature fusion across the extended spatial hierarchy required for outdoor registration.

\begin{table}
\centering
\caption{KPConv-FPN architecture for KITTI. SAGE uses a 5-stage backbone with an additional downsampling stage (Stage 5) compared to 3DMatch to handle larger outdoor scenes. The input feature dimension of the first KPConv layer is modified from 1 to 1088 (highlighted in red) to incorporate semantic features.}
\label{tab:architecture_kitti}
\setlength{\tabcolsep}{4.2pt}
\resizebox{\linewidth}{!}{
\begin{tabular}{c|c|c}
\toprule
Stage & \multicolumn{1}{c|}{GeoTransformer} & \multicolumn{1}{c}{SAGE} \\
\midrule
\multicolumn{3}{c}{\textit{Downsampling}} \\
\midrule
1 & KPConv($1 \rightarrow 64$) & KPConv(\textcolor{red}{1088} $\rightarrow 64$) \\
  & ResBlock($64 \rightarrow 128$) & ResBlock($64 \rightarrow 128$) \\
\midrule
2 & ResBlock($128 \rightarrow 128$, strided) & ResBlock($128 \rightarrow 128$, strided) \\
  & ResBlock($128 \rightarrow 256$) & ResBlock($128 \rightarrow 256$) \\
  & ResBlock($256 \rightarrow 256$) & ResBlock($256 \rightarrow 256$) \\
\midrule
3 & ResBlock($256 \rightarrow 256$, strided) & ResBlock($256 \rightarrow 256$, strided) \\
  & ResBlock($256 \rightarrow 512$) & ResBlock($256 \rightarrow 512$) \\
  & ResBlock($512 \rightarrow 512$) & ResBlock($512 \rightarrow 512$) \\
\midrule
4 & ResBlock($512 \rightarrow 512$, strided) & ResBlock($512 \rightarrow 512$, strided) \\
  & ResBlock($512 \rightarrow 1024$) & ResBlock($512 \rightarrow 1024$) \\
  & ResBlock($1024 \rightarrow 1024$) & ResBlock($1024 \rightarrow 1024$) \\
\midrule
5 & ResBlock($1024 \rightarrow 1024$, strided) & ResBlock($1024 \rightarrow 1024$, strided) \\
  & ResBlock($1024 \rightarrow 2048$) & ResBlock($1024 \rightarrow 2048$) \\
  & ResBlock($2048 \rightarrow 2048$) & ResBlock($2048 \rightarrow 2048$) \\
\midrule
\multicolumn{3}{c}{\textit{Upsampling (FPN)}} \\
\midrule
6 & NearestUpsampling & NearestUpsampling \\
  & UnaryConv($3072 \rightarrow 1024$) & UnaryConv($3072 \rightarrow 1024$) \\
\midrule
7 & NearestUpsampling & NearestUpsampling \\
  & UnaryConv($1536 \rightarrow 512$) & UnaryConv($1536 \rightarrow 512$) \\
\midrule
8 & NearestUpsampling & NearestUpsampling \\
  & UnaryConv($768 \rightarrow 256$) & UnaryConv($768 \rightarrow 256$) \\
\bottomrule
\end{tabular}
}

\end{table}

\subsection{Semantic-Guided MoE Transformer}
\label{sec:a_sgmoe}
This subsection details the architecture of the SG-MoE Transformer introduced in Sec. 3.4 of the main paper, which refines the superpoint-level semantic-geometric features before matching.
The module operates on $d_{sg} = 256$-dimensional semantic-geometric features and stacks three blocks, each consisting of a geometric self-attention layer followed by an SG-MoE cross-attention layer (i.e., the block pattern [self, cross] repeated three times). The cross-attention layer uses $4$ heads.

\noindent\textbf{Router.}
Routing is computed from the superpoint-level frozen semantic features $\hat{\mathbf{f}}_{s,k} \in \mathbb{R}^{1024}$ (Sonata~\cite{wu2025sonata} output, aggregated to the superpoint level as described in Sec. 3.3 of the main paper). The routing network $\pi$ (Eq.~(2) of the main paper) is a two-layer MLP ($1024 \rightarrow 256 \rightarrow N_{\text{expert}}$) with ReLU activation, followed by a softmax and top-$1$ selection. We apply a gate dropout of $0.1$.

\noindent\textbf{Experts.}
Each of the $N_{\text{expert}} = 4$ experts is a two-layer feed-forward network with ReLU activation and an inverted-bottleneck shape ($256 \rightarrow 512 \rightarrow 256$), with an expert dropout of $0.1$. The experts process the semantic-geometric features produced by cross-attention, not the semantic features used for routing.

\subsection{Superpoint Matching Module}
\label{sec:a_3}

Following GeoTransformer~\cite{qin2022geotransformer}, we adopt their superpoint matching strategy to establish coarse correspondences between downsampled point clouds. The module consists of three key operations: Gaussian correlation computation, dual-normalization, and top-k correspondence selection.

\noindent\textbf{Gaussian Correlation Matrix.}
Given the refined semantic-geometric features $\hat{\mathbf{F}}^{\mathcal{P}}_{\text{out}} \in \mathbb{R}^{\hat{N}_{\mathcal{P}} \times d_{sg}}$ and $\hat{\mathbf{F}}^{\mathcal{Q}}_{\text{out}} \in \mathbb{R}^{\hat{N}_{\mathcal{Q}} \times d_{sg}}$ from the SG-MoE Transformer, we first normalize them onto a unit hypersphere and compute a Gaussian correlation matrix $\mathbf{S} \in \mathbb{R}^{\hat{N}_{\mathcal{P}} \times \hat{N}_{\mathcal{Q}}}$:
\begin{equation}
s_{i,j} = \exp\left(-\|\hat{\mathbf{f}}^{\mathcal{P}}_{\text{out},i} - \hat{\mathbf{f}}^{\mathcal{Q}}_{\text{out},j}\|^2_2\right),
\end{equation}
where $\hat{\mathbf{f}}^{\mathcal{P}}_{\text{out},i}$ and $\hat{\mathbf{f}}^{\mathcal{Q}}_{\text{out},j}$ are the $i$-th and $j$-th rows of normalized feature matrices.

\noindent\textbf{Dual-normalization.}
To suppress ambiguous matches from geometrically indiscriminative patches, we apply a dual-normalization operation~\cite{rocco2018neighbourhood,sun2021loftr} on the correlation matrix:
\begin{equation}
\bar{s}_{i,j} = \frac{s_{i,j}}{\sum_{k=1}^{\hat{N}_{\mathcal{Q}}} s_{i,k}} \cdot \frac{s_{i,j}}{\sum_{k=1}^{\hat{N}_{\mathcal{P}}} s_{k,j}}.
\end{equation}
This operation normalizes the correlation scores both row-wise and column-wise, effectively penalizing matches where a superpoint has multiple high-scoring candidates.

\noindent\textbf{Top-k Correspondence Selection.}
We select the $N_c$ highest-scoring entries from the normalized correlation matrix $\bar{\mathbf{S}}$ as the superpoint correspondences:
\begin{equation}
\hat{\mathcal{C}} = \{(\hat{\mathbf{p}}_{x_i}, \hat{\mathbf{q}}_{y_i}) \mid (x_i, y_i) \in \text{topk}_{x,y}(\bar{s}_{x,y})\}.
\end{equation}
Following GeoTransformer~\cite{qin2022geotransformer}, we use $N_c = 256$ putative superpoint matches during inference.

\noindent\textbf{Empty Patch Removal.}
Before matching, we remove superpoints with empty local patches. A superpoint $\hat{\mathbf{p}}_i$ is considered empty if its associated patch $\mathcal{G}^{\mathcal{P}}_i$ (i.e., the set of points assigned to $\hat{\mathbf{p}}_i$ by the point-to-node grouping of GeoTransformer~\cite{qin2022geotransformer}) contains no points. This preprocessing step improves matching quality by excluding uninformative superpoints.

\subsection{Point Matching Module}
\label{sec:a_4}
For each superpoint correspondence $(\hat{\mathbf{p}}_{x_i}, \hat{\mathbf{q}}_{y_i}) \in \hat{\mathcal{C}}$, we extract dense point correspondences from their local patches using an optimal transport layer~\cite{sarlin2020superglue}. 

Following the same procedure as in GeoTransformer~\cite{qin2022geotransformer}, we compute a cost matrix $\mathbf{C}_i \in \mathbb{R}^{n_i \times m_i}$ from intermediate-resolution point features $\mathbf{F}^{\mathcal{P}}_{x_i}$ and $\mathbf{F}^{\mathcal{Q}}_{y_i}$:
\begin{equation}
\mathbf{C}_i = \mathbf{F}^{\mathcal{P}}_{x_i}(\mathbf{F}^{\mathcal{Q}}_{y_i})^T / \sqrt{\tilde{d}},
\end{equation}
where $n_i = |\mathcal{G}^{\mathcal{P}}_{x_i}|$, $m_i = |\mathcal{G}^{\mathcal{Q}}_{y_i}|$ are patch sizes, and $\tilde{d}$ is the feature dimension.

This cost matrix is augmented to $\bar{\mathbf{C}}_i \in \mathbb{R}^{(n_i+1) \times (m_i+1)}$ by appending dustbin entries (initialized with learnable parameter $\alpha$) to handle unmatched points. The Sinkhorn algorithm~\cite{sinkhorn1967concerning} then produces soft assignments $\mathbf{A}_i$, from which we extract correspondences via mutual top-$k$ selection after removing dustbin entries. We set $k=3$ to obtain approximately 6000 correspondences for Local-to-Global Registration.

\subsection{Local-to-Global Registration}
\label{sec:a_5}
We adopt GeoTransformer's Local-to-Global Registration (LGR)~\cite{qin2022geotransformer}, a parameter-free and RANSAC-free method. 

\textbf{Local Phase:} For each superpoint correspondence, we solve for transformation $\mathbf{T}_i = (\mathbf{R}_i, \mathbf{t}_i)$ using weighted SVD over local point correspondences $\mathcal{C}_i$:
\begin{equation}
\mathbf{R}_i, \mathbf{t}_i = \arg\min_{\mathbf{R}, \mathbf{t}} \sum_{(\tilde{\mathbf{p}}_{x_j}, \tilde{\mathbf{q}}_{y_j}) \in \mathcal{C}_i} w^i_j \|\mathbf{R} \cdot \tilde{\mathbf{p}}_{x_j} + \mathbf{t} - \tilde{\mathbf{q}}_{y_j}\|^2_2,
\end{equation}
where weights $w^i_j$ are confidence scores from the optimal transport layer.

\textbf{Global Phase:} Select the transformation admitting the most inliers:
\begin{equation}
\scalebox{0.85}{$\displaystyle
\mathbf{R}, \mathbf{t} = \arg\max_{\mathbf{R}_i, \mathbf{t}_i} \sum_{(\tilde{\mathbf{p}}_{x_j}, \tilde{\mathbf{q}}_{y_j}) \in \mathcal{C}} \mathbb{I}[\|\mathbf{R}_i \cdot \tilde{\mathbf{p}}_{x_j} + \mathbf{t}_i - \tilde{\mathbf{q}}_{y_j}\|_2 < \tau_a],$}
\end{equation}
with acceptance radius $\tau_a = 0.1$\,m. We refine this transformation iteratively for $N_r = 5$ iterations.

\section{Loss Function Details}
\label{sec:app_loss}

\subsection{Overlap-Aware Circle Loss}
\label{sec:b_1}
We supervise superpoint matching with an overlap-aware circle loss adapted from GeoTransformer~\cite{qin2022geotransformer}. For anchor superpoints $\mathcal{A}^{\mathcal{P}}$ (those with at least one overlapping patch in $\hat{\mathcal{Q}}$), we define:
\begin{itemize}
    \item $\mathcal{P}_k^{+}$: positive patches (overlap $\geq 10\%$)
    \item $\mathcal{P}_k^{-}$: negative patches (no overlap)
\end{itemize}

The loss for the direction $\mathcal{P} \to \mathcal{Q}$ is:
\begin{equation}
\scalebox{0.85}{$\displaystyle
\mathcal{L}_{\text{oc}}^{\mathcal{P} \to \mathcal{Q}} = \frac{1}{|\mathcal{A}^{\mathcal{P}}|} \sum_{k \in \mathcal{A}^{\mathcal{P}}} \log\left(1 + \sum_{j \in \mathcal{P}_k^{+}} e^{w_{k,j} \beta_{k,j}^{+}} \cdot \sum_{\ell \in \mathcal{P}_k^{-}} e^{\beta_{k,\ell}^{-}}\right),$}
\end{equation}
where:
\begin{itemize}
    \item $d_{k,j} = \|\hat{\mathbf{f}}_{\text{out},k}^{\mathcal{P}} - \hat{\mathbf{f}}_{\text{out},j}^{\mathcal{Q}}\|_2$ (distance in the refined semantic-geometric feature space)
    \item $w_{k,j} = \sqrt{o_{k,j}}$ (overlap-based weighting)
    \item $\beta_{k,j}^{+} = \gamma(d_{k,j} - \Delta_p)$ and $\beta_{k,\ell}^{-} = \gamma(\Delta_n - d_{k,\ell})$
\end{itemize}

The bidirectional coarse loss is $\mathcal{L}_{\text{coarse}} = \mathcal{L}_{\text{oc}}^{\mathcal{P} \to \mathcal{Q}} + \mathcal{L}_{\text{oc}}^{\mathcal{Q} \to \mathcal{P}}$.

\subsection{Optimal Transport Loss}
\label{sec:b_2}
For each sampled ground-truth superpoint pair during training, we supervise point-level matches via negative log-likelihood loss~\cite{sarlin2020superglue}. 

Given cost matrix $\mathbf{C}_i$ augmented to $\bar{\mathbf{C}}_i$ with dustbin entries, Sinkhorn iterations produce soft assignment $\mathbf{A}_i$. Ground-truth labels $\Gamma_i$ include:
\begin{itemize}
    \item Matched pairs: $(p, q)$ with $\|\mathbf{T}^*(p) - q\|_2 < 0.05$\,m
    \item Dustbin assignments for unmatched points
\end{itemize}

The fine loss is:
\begin{equation}
\mathcal{L}_{\text{fine}} = \frac{1}{N_g} \sum_{i=1}^{N_g} \left( -\sum_{(p,q) \in \Gamma_i} \log a_{p,q}^i \right),
\end{equation}
where we sample $N_g=128$ ground-truth superpoint pairs per batch.

\subsection{Expert Load Balancing Loss}
\label{sec:b_3}
The SG-MoE layer introduces a set of trainable expert networks, which necessitates an auxiliary loss to ensure training stability and prevent expert saturation. 
This is a common failure mode in MoE models where the router learns to favor only one or a few popular experts, leaving the others untrained and wasting capacity.
To enforce balanced expert utilization, we employ a load balancing loss $\mathcal{L}_{\text{moe}}$~\cite{shazeer2017outrageously}.

This loss encourages the routing network to distribute the superpoints uniformly across all $N_{\text{expert}}$ experts on average over the batch.
The routing network computes a probability distribution $\mathbf{r}_k = [r_{k,1}, \ldots, r_{k,N_{\text{expert}}}]$ for each superpoint $k$.
We first compute the average routing probability for each expert $e$ by averaging its probability $r_{k,e}$ across all superpoints in the batch $\mathcal{B}$ (including superpoints from both $\mathcal{P}$ and $\mathcal{Q}$):
\begin{equation}
\bar{r}_e = \frac{1}{|\mathcal{B}|} \sum_{k \in \mathcal{B}} r_{k,e},
\end{equation}
where $|\mathcal{B}|$ is the total number of superpoints in the batch.
This gives us the batch-wise load distribution $\bar{\mathbf{r}} = [\bar{r}_1, \ldots, \bar{r}_{N_{\text{expert}}}]$.
We want this distribution $\bar{\mathbf{r}}$ to be as close as possible to a uniform distribution $\mathbf{u} = [1/N_{\text{expert}}, \ldots, 1/N_{\text{expert}}]$. 
We achieve this by minimizing the Kullback-Leibler (KL) divergence between $\bar{\mathbf{r}}$ and $\mathbf{u}$:
\begin{equation}
\mathcal{L}_{\text{moe}} = KL(\bar{\mathbf{r}} || \mathbf{u}) = \sum_{e=1}^{N_{\text{expert}}} \bar{r}_e \log \frac{\bar{r}_e}{1/N_{\text{expert}}}.
\end{equation}
Minimizing this loss penalizes routing strategies that deviate significantly from uniformity, encouraging the model to leverage the full capacity of all $N_{\text{expert}}$ specialized experts.

\subsection{Training Configuration}
\label{sec:b_4}
\textbf{Optimizer:} Adam with learning rate $10^{-4}$, multiplied by 0.95 after each epoch, for 45 epochs.

\textbf{Loss hyperparameters:}
\begin{itemize}
    \item Coarse loss: $\Delta_p = 0.1$, $\Delta_n = 1.4$, $\gamma = 80$
    \item Fine loss: matching radius $\tau = 0.05$\,m
    \item MoE load balancing: $\lambda_{\text{moe}} = 0.01$
\end{itemize}

\textbf{Random seed:} 7351

\section{Evaluation Metrics}
\label{sec:app_metrics}

We provide detailed mathematical definitions of the evaluation metrics used in our experiments.
The metrics differ between indoor (3DMatch/3DLoMatch) and outdoor (KITTI) benchmarks due to different evaluation protocols and scene characteristics.

\subsection{3DMatch and 3DLoMatch}
\label{sec:c_1}
\noindent\textbf{Inlier Ratio (IR).}
The Inlier Ratio measures the quality of correspondences as the fraction of inlier matches among all putative correspondences:
\begin{equation}
    \text{IR} = \frac{1}{|\mathcal{C}|}\sum_{(p_i, q_j) \in \mathcal{C}} \mathbb{I}[\|T^*(p_i) - q_j\|_2 < \tau_{\text{inlier}}],
\end{equation}
where $\mathcal{C}$ denotes the set of putative correspondences, $T^*$ is the ground-truth transformation, $\mathbb{I}[\cdot]$ is the indicator function, and $\tau_{\text{inlier}} = 0.1$\,m is the inlier threshold.

\noindent\textbf{Feature Matching Recall (FMR).}
The Feature Matching Recall evaluates the percentage of point cloud pairs that have sufficient inliers for potential successful registration:
\begin{equation}
    \text{FMR} = \frac{1}{N}\sum_{i=1}^{N} \mathbb{I}[\text{IR}_i > \tau_{\text{FMR}}],
\end{equation}
where $N$ is the total number of point cloud pairs in the dataset, and $\tau_{\text{FMR}} = 0.05$ (5\%).

\noindent\textbf{Registration Recall (RR).}
The Registration Recall measures the fraction of successfully registered point cloud pairs:
\begin{equation}
    \text{RR} = \frac{1}{N}\sum_{i=1}^{N} \mathbb{I}[\text{RMSE}_i < \tau_{\text{RR}}],
\end{equation}
where the RMSE (Root Mean Square Error) is computed over ground-truth correspondences $\mathcal{C}^*$ under the estimated transformation $\hat{T}$:
\begin{equation}
    \text{RMSE} = \sqrt{\frac{1}{|\mathcal{C}^*|}\sum_{(p_i^*, q_j^*) \in \mathcal{C}^*} \|\hat{T}(p_i^*) - q_j^*\|_2^2},
\end{equation}
and $\tau_{\text{RR}} = 0.2$\,m is the success threshold.

\subsection{KITTI}
\label{sec:c_2}
For outdoor KITTI odometry benchmark, we adopt different metrics that focus on the accuracy of the estimated transformation itself rather than correspondence quality.

\noindent\textbf{Relative Translation Error (RTE).}
The Relative Translation Error measures the Euclidean distance between the estimated translation vector and the ground-truth translation vector:
\begin{equation}
    \text{RTE} = \|\hat{\mathbf{t}} - \mathbf{t}^*\|_2,
\end{equation}
where $\hat{\mathbf{t}}$ is the estimated translation and $\mathbf{t}^*$ is the ground-truth translation.
We report RTE in centimeters.

\noindent\textbf{Relative Rotation Error (RRE).}
The Relative Rotation Error measures the geodesic distance between the estimated rotation matrix and the ground-truth rotation matrix:
\begin{equation}
    \text{RRE} = \arccos\left(\frac{\text{trace}(\hat{\mathbf{R}}^T \mathbf{R}^*) - 1}{2}\right),
\end{equation}
where $\hat{\mathbf{R}}$ is the estimated rotation matrix and $\mathbf{R}^*$ is the ground-truth rotation matrix.
We report RRE in degrees.
The geodesic distance represents the angle of rotation needed to align $\hat{\mathbf{R}}$ with $\mathbf{R}^*$.

\noindent\textbf{Registration Recall (RR).}
Following the KITTI evaluation protocol~\cite{geiger2012kitti}, the Registration Recall measures the fraction of point cloud pairs whose transformation errors are below specified thresholds:
\begin{equation}
    \text{RR} = \frac{1}{N}\sum_{i=1}^{N} \mathbb{I}[\text{RTE}_i < \tau_t \land \text{RRE}_i < \tau_r],
\end{equation}
where $\tau_t = 2$\,m is the translation error threshold, $\tau_r = 5^\circ$ is the rotation error threshold, and $N$ is the total number of point cloud pairs.
Note that both conditions must be satisfied simultaneously for a pair to be considered successfully registered.

\noindent\textbf{Mean error computation.}
Following standard practice~\cite{qin2022geotransformer,huang2021predator}, when reporting mean RTE and RRE values, we compute the average only over successfully registered pairs (those satisfying the RR criteria) to avoid including catastrophic failures that would dominate the statistics.

\section{Additional Experiments}
\label{sec:app_add}
Unless otherwise stated, all experiments in this section employ Local-to-Global Registration (LGR)~\cite{qin2022geotransformer} as the transformation estimator, which is a RANSAC-free method that leverages confidence-weighted correspondences for robust pose estimation.
As discussed in the main paper, LGR is better suited to exploit the semantically structured correspondences produced by SAGE, and therefore serves as the default estimator throughout our supplementary evaluation.

\subsection{Outdoor Scene Performance on KITTI}
To evaluate the generalization capability of SAGE to outdoor environments, we conduct experiments on the KITTI odometry benchmark~\cite{geiger2012kitti}. 
Following the standard protocol, we measure Relative Translation Error (RTE), Relative Rotation Error (RRE), and Registration Recall (RR) as defined in Sec.~\ref{sec:c_2}.

As shown in Table~\ref{tab:kitti}, SAGE achieves competitive performance with RTE of 6.1\,cm, RRE of 0.22$^\circ$, and RR of 99.8\%, demonstrating that our semantic-guided approach generalizes well to outdoor large-scale scenarios.
The results confirm that semantic representations learned from indoor scenes transfer effectively to outdoor environments, validating the robustness of our method across diverse scene types.

\begin{table}[t]
\centering
\begin{tabular}{c|c c c}
\toprule
    &     RTE[cm]$\downarrow$ &	RRE[$^\circ$]$\downarrow$ &    RR[\%]$\uparrow$   \\
\midrule
FCGF      &     9.5     &	0.3	    &    96.6 \\
D3Feat    &     7.2     &	0.3	    &    99.8 \\
FMR       &     $\sim$66     &	1.49	&    90.6 \\    
DGR       &     $\sim$32     &	0.37	&    98.7 \\    
SpinNet   &     9.9     &	0.47	&    99.1 \\    
Predator  &     6.8     &	0.27	&    99.8 \\    
HregNet   &     $\sim$12     &	0.29	&    99.7 \\    
CoFiNet   &     8.2     &	0.41	&    99.8 \\    
GeoTrans  &     6.8     &	0.24	&    99.8 \\    
DCATr     &     6.6     &	0.22	&    99.7 \\    
SAGE      &     6.1     &	0.22	&    99.8 \\    
\bottomrule
\end{tabular}
\caption{Registration performance on KITTI.}
\label{tab:kitti}
\end{table}

\subsection{Scene-wise Analysis}
\label{sec:d_2_1}

To evaluate the generalization capability of SAGE across diverse environments, we conduct a scene-wise analysis spanning both the 3DMatch and 3DLoMatch benchmarks.
Since these two benchmarks share the same set of test scenes but differ in the overlap ratio threshold used for selecting point cloud pairs, comparing results across both settings allows us to examine whether the benefit of semantic guidance is consistent regardless of the degree of overlap.
This analysis also serves to clarify when and why the SG-MoE routing mechanism provides the most substantial gains.

\begin{table*}
\centering

\setlength{\tabcolsep}{4.2pt}
\resizebox{0.8\linewidth}{!}{
\begin{tabular}{l|cccccccc|c}
\toprule
Scenes & \textit{Kitchen} & \textit{Home 1} & \textit{Home 2} & \textit{Hotel 1} & \textit{Hotel 2} & \textit{Hotel 3} & \textit{Study} & \textit{MIT Lab} & \textbf{All} \\ \midrule
\#Pairs & 1031 & 445 & 438 & 444 & 262 & 103 & 532 & 149 & 3404 \\ \midrule
GeoTrans. & 91.7 & \textbf{84.9} & \underline{85.4} & \underline{94.1} & 80.2 & \textbf{86.4} & \underline{75.2} & \textbf{83.9} & 86.3 \\
RoITr & \underline{94.0} & 81.6 & \underline{85.4} & 93.9 & \underline{80.9} & \underline{85.4} & 74.3 & \underline{81.9} & \underline{86.4} \\
\textbf{SAGE} (Ours) & \textbf{94.2} & \underline{82.3} & \textbf{86.1} & \textbf{95.7} & \textbf{82.8} & \underline{85.4} & \textbf{80.3} & 81.2 & \textbf{87.9} \\ \bottomrule
\end{tabular}
}
\caption{Registration Recall (\%) on the 3DMatch and 3DLoMatch benchmark across different scene categories. \#Pairs denotes the total number of point cloud pairs evaluated in each scene.}
\label{tab:scene_results}
\end{table*}

Table~\ref{tab:scene_results} summarizes the Registration Recall (RR) for eight distinct scenes. We compare SAGE against GeoTransformer and RoITr.

The results demonstrate that SAGE achieves the highest overall RR of 87.9\%, improving over RoITr (86.4\%) and GeoTransformer (86.3\%).
A closer look at individual scenes reveals a clear pattern that is closely tied to the semantic diversity of each environment.
SAGE achieves the largest gains in scenes that contain a rich variety of semantically distinct objects.
For instance, in the \textit{Study} scene, which typically contains diverse cluttered objects such as desks, bookshelves, monitors, and chairs, SAGE achieves 80.3\% RR, significantly outperforming GeoTransformer (75.2\%) and RoITr (74.3\%) by over 5 percentage points.
This substantial improvement can be attributed to the SG-MoE router, which assigns superpoints belonging to different object categories to specialized experts that can extract discriminative semantic-geometric representations for each category.
In scenes with diverse objects, this expert specialization provides clear advantages because each expert can learn category-specific feature refinement strategies rather than relying on a single shared transformation.
Similarly, in the \textit{Kitchen} scene, which contains a variety of appliances, counters, and cabinets that are geometrically similar but semantically different, SAGE achieves 94.2\% RR compared to GeoTransformer's 91.7\%.

On the other hand, SAGE shows relatively smaller gains or comparable performance in scenes dominated by semantically homogeneous structures.
In the \textit{Hotel 3} scene, all three methods achieve similar results (85.4--86.4\%), because the scene is largely composed of uniform corridors and walls that offer limited semantic diversity for the SG-MoE router to exploit.
In the \textit{MIT Lab} scene, SAGE (81.2\%) slightly underperforms GeoTransformer (83.9\%).
This scene is characterized by long, narrow laboratory corridors with repetitive structural elements, where the overlapping regions are dominated by flat walls and floors that provide minimal semantic variation.

\subsection{Overlap-wise Analysis}
\label{sec:d_2_2}

In real-world scenarios, point cloud pairs often exhibit low overlapping areas, which significantly increases the difficulty of correspondence search. To assess the robustness of SAGE against such partiality, we evaluate the performance as a function of the overlap ratio.

\begin{table*}
\centering
\setlength{\tabcolsep}{4.2pt}
\resizebox{0.8\linewidth}{!}{
\begin{tabular}{l|cccccccccc}
\toprule
Overlap Ratio & 0-10\% & 10-20\% & 20-30\% & 30-40\% & 40-50\% & 50-60\% & 60-70\% & 70-80\% & 80-90\% & 90-100\% \\ \midrule
\#Pairs & 3 & 1040 & 832 & 387 & 417 & 304 & 224 & 131 & 58 & 8 \\ \midrule
GeoTrans. & \textbf{66.7} & 70.6 & \underline{86.4} & \underline{94.8} & \underline{95.2} & \textbf{98.7} & \textbf{99.6} & \textbf{100.0} & \textbf{100.0} & \textbf{100.0} \\
RoITr & \textbf{66.7} & \underline{72.2} & 85.5 & 94.6 & 94.5 & 98.0 & 98.7 & \textbf{100.0} & \textbf{100.0} & \textbf{100.0} \\
\textbf{SAGE} (Ours) & \textbf{66.7} & \textbf{73.4} & \textbf{88.7} & \textbf{95.3} & \textbf{96.2} & \underline{98.4} & \textbf{99.6} & \textbf{100.0} & \textbf{100.0} & \textbf{100.0} \\ \bottomrule
\end{tabular}
}
\caption{Registration Recall (\%) comparison across different overlap ratios. \#Pairs denotes the number of point cloud pairs in each overlap bin.}
\label{tab:overlap_results}
\end{table*}

We categorize the test pairs into ten bins based on their ground-truth overlap ratios, ranging from 0\% to 100\% with a bin width of 10\%. The RR for each bin is reported in Table~\ref{tab:overlap_results}.

As shown in Table~\ref{tab:overlap_results}, SAGE demonstrates a clear and consistent advantage across the low-to-moderate overlap regime (10--50\%), which represents the most challenging setting for correspondence-based registration.
In the 10--20\% bin, which contains the largest number of test pairs (1,040 pairs) and poses the greatest challenge, SAGE achieves 73.4\% RR, outperforming RoITr (72.2\%) and GeoTransformer (70.6\%) by 1.2 and 2.8 percentage points, respectively.
In the 20--30\% bin, the improvement over RoITr is even more pronounced: SAGE reaches 88.7\% RR compared to RoITr's 85.5\% (+3.2 points) and GeoTransformer's 86.4\% (+2.3 points).
This pattern extends through the 30--50\% range, where SAGE consistently achieves the highest RR.
These results provide direct evidence for the effectiveness of the SG-MoE routing mechanism.
In low-overlap scenarios, the visible portion of each point cloud is limited, and the few geometric structures available for matching are often ambiguous.
The SG-MoE router addresses this by leveraging semantic cues to assign each superpoint to a specialized expert, enabling the model to distinguish between geometrically similar but semantically different structures even when geometric context alone is insufficient.

In the 50--60\% bin, SAGE (98.4\%) is slightly below GeoTransformer (98.7\%). At this overlap level, the registration problem becomes substantially easier because ample geometric context is available, and the benefit of semantic guidance diminishes accordingly. This marginal difference is consistent with our observation from the scene-wise analysis that semantic routing is most valuable when geometric cues alone are insufficient.

The relationship between overlap ratio and the relative advantage of SAGE further validates that the SG-MoE architecture is well suited for the most practically important operating regime, namely scenarios with limited overlap, which frequently arise in real-world applications such as large-scale SLAM and autonomous navigation.

\subsection{Computational Efficiency}
\label{sec:d_2}
Table~\ref{tab:computational_efficiency_analysis} shows that SAGE requires approximately 12.4$\times$ more model parameters and 2.4$\times$ more memory than GeoTransformer, due to introducing Sonata and MoE.
While inference time increases to 179ms, this overhead is comparable to RoITr (140ms), which also employs sophisticated architectures for improved performance.

\begin{table}

\centering
\setlength{\tabcolsep}{4.2pt}
\resizebox{\linewidth}{!}{
\begin{tabular}{l|cccc}
\toprule
\textbf{Method} & Params (M) & Mem. (GB)* & Time (ms)** & RR (\%) \\
\midrule
GeoTransformer & 9.83   & 7.02 & 75 & 74.0  \\
RoITr          & 10.10  & 9.28 & 140 & 73.4  \\
\textbf{SAGE} (Ours)  & 121.71 & 16.69 & 179 & 76.3  \\
\bottomrule
\end{tabular}
}
\\
\vspace{4pt}
* Memory measured using PyTorch memlab.\\
** Inference time measured on NVIDIA H100.
\caption{Computational Efficiency Analysis}
\label{tab:computational_efficiency_analysis}
\end{table}

SAGE has 121.71M total parameters, which is approximately 12.4 times larger than GeoTransformer (9.83M).
However, the majority of this increase originates from the frozen Sonata encoder (108.41M parameters), which requires no gradient computation during training or inference.
Excluding Sonata, SAGE contains only 13.3M parameters; the SG-MoE routing networks and expert FFNs add approximately 3.5M parameters beyond the GeoTransformer backbone, representing only a 35\% increase.

Regarding inference time, SAGE requires 179ms per point cloud pair, comparable to RoITr (140ms).
In terms of performance gain per unit of additional latency over GeoTransformer, SAGE achieves 2.3 percentage points of RR improvement for 104ms of additional latency, whereas RoITr adds 65ms without improving RR (73.4\% vs. 74.0\%), indicating that SAGE utilizes the additional computation more effectively.

It is worth noting that this computational overhead is not specific to SAGE, but rather reflects a fundamental cost of incorporating pre-trained semantic encoders into the registration pipeline.
SAGE mitigates this inherent cost by freezing the encoder entirely and keeping the method-specific trainable overhead minimal.
The peak memory consumption of 16.69 GB fits within commonly available GPUs such as the NVIDIA RTX 4090 (24 GB), making SAGE practical for offline applications that prioritize accuracy, such as 3D reconstruction for cultural heritage preservation and architectural surveying.
As more lightweight point cloud foundation models emerge, the frozen encoder could be replaced with a more compact alternative without modifying the SG-MoE architecture itself.

\subsection{Additional Ablation Study}
\label{sec:d_3}
We conduct additional ablation studies to further validate the key design choices in SAGE.
Specifically, we investigate three critical questions: (1) \textit{Should the semantic encoder be adapted during training?} We examine whether fine-tuning Sonata with adapters improves performance or causes geometric shortcuts. (2) \textit{How should semantic features guide expert routing?} We compare our query-side routing against cross-attention-based alternatives. (3) \textit{How many experts are optimal?} We vary the number of experts from 2 to 6 to understand the capacity-efficiency trade-off.
All experiments are conducted on 3DLoMatch with LGR, and results are summarized in Table~\ref{tab:ablation_additional}.

\begin{table}

\centering
\setlength{\tabcolsep}{4.2pt}
\begin{tabular}{l|ccc}
\toprule
\textbf{Configuration} & \textbf{FMR} & \textbf{IR} & \textbf{RR} \\
\midrule
\multicolumn{4}{c}{\textit{Sonata Adaptation}} \\
\midrule
w/o adapter & 89.4 & 44.3 & 76.3 \\
w/ adapter   & 86.4 & 42.6 & 74.8 \\
\midrule
\multicolumn{4}{c}{\textit{Routing Strategy}} \\
\midrule
SG-MoE (ours) & 89.4 & 44.3 & 76.3 \\
SCAR          & 88.1 & 44.4 & 75.0 \\
\midrule
\multicolumn{4}{c}{\textit{Number of Experts}} \\
\midrule
$N_\text{expert} = 2$ & 88.0 & 42.3 & 74.9 \\
$N_\text{expert} = 3$ & 89.3 & 41.8 & 75.1 \\
$N_\text{expert} = 4$ & 89.4 & 44.3 & 76.3 \\
$N_\text{expert} = 5$ & 88.5 & 43.6 & 75.5 \\
$N_\text{expert} = 6$ & 88.9 & 43.0 & 75.7 \\
\bottomrule
\end{tabular}
\caption{Additional ablation studies on 3DLoMatch with LGR. All metrics in \%. We evaluate: (1) adding adapters to Sonata, (2) alternative routing strategy using cross-attention (SCAR), and (3) varying the number of experts.}
\label{tab:ablation_additional}
\end{table}

\noindent\textbf{Sonata with Adapter.}
As shown in Table~\ref{tab:ablation_additional}(top), adding adapters to Sonata (w/ adapter) achieves 74.8\% RR, showing a notable degradation ($-1.5$ points) compared to the fully frozen Sonata baseline (76.3\% RR).
This result demonstrates that allowing adaptation of the semantic encoder, even with lightweight adapter modules, causes the model to fall into a \textit{geometric shortcut}~\cite{wu2025sonata}.
Specifically, the registration loss functions, which optimize primarily for geometric alignment accuracy, drive the adapted semantic features to collapse toward purely geometric patterns that are more directly useful for minimizing registration errors.
This geometric shortcut phenomenon causes the semantic representations to lose their ability to distinguish between geometrically similar but semantically different structures (e.g., TV corners vs. desk corners), which is precisely the discriminative power that Sonata was pre-trained to preserve.

The degradation in FMR (89.4\% $\rightarrow$ 86.4\%) and IR (44.3\% $\rightarrow$ 42.6\%) further confirms that the adapted features become less effective at establishing reliable correspondences in low-overlap scenarios where semantic understanding is crucial.
This validates our design choice of keeping Sonata completely frozen throughout training, ensuring that the semantic representations maintain their original semantic discriminability and do not degenerate into redundant geometric features.

\noindent\textbf{Alternative Routing Strategy: SCAR.}
\begin{figure}[t]
    \centering
    \includegraphics[width=0.85\linewidth]{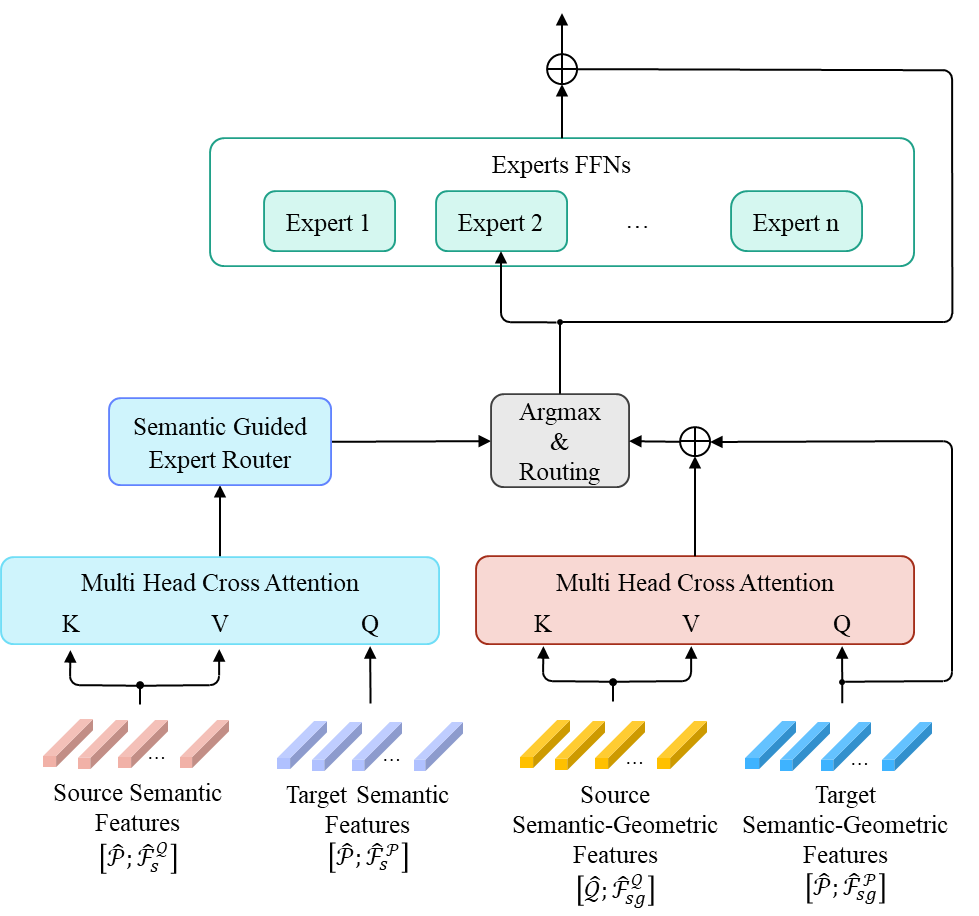}
    \caption{
    \textbf{The Overview of SCAR.} 
    Semantic Cross-Attention Routing (SCAR) first computes cross-attention between source and target semantic features, then uses the attended features for routing.
    While SCAR incorporates inter-point-cloud information earlier, this premature soft matching introduces noise in low-overlap scenarios, leading to inferior performance (76.3\% vs. 75.0\% RR on 3DLoMatch).
    }
    \label{fig:scar}
\end{figure}
We explore an alternative routing mechanism called Semantic Cross-Attention Routing (SCAR), where expert selection is determined by cross-attention between source and target semantic features rather than using only the query point cloud's semantic features.

As illustrated in Fig.~\ref{fig:scar}, SCAR computes routing probabilities by first performing cross-attention between the semantic features of the two point clouds.
Specifically, for each superpoint in $\mathcal{P}$, we compute cross-attention scores with all superpoints in $\mathcal{Q}$ using their frozen semantic representations $\hat{\mathbf{F}}_s^{\mathcal{P}}$ and $\hat{\mathbf{F}}_s^{\mathcal{Q}}$:
\begin{equation}
\mathbf{A}_{i,j} = \text{Softmax}\left(\frac{\hat{\mathbf{f}}_{s,i}^{\mathcal{P}} (\hat{\mathbf{f}}_{s,j}^{\mathcal{Q}})^T}{\sqrt{d_s}}\right),
\end{equation}
and then aggregate the attended features to compute routing probabilities:
\begin{equation}
\mathbf{s}_i^{\text{SCAR}} = \text{Softmax}\left(\text{MLP}_r\left(\sum_{j=1}^{\hat{N}_{\mathcal{Q}}} \mathbf{A}_{i,j} \cdot \hat{\mathbf{f}}_{s,j}^{\mathcal{Q}}\right)\right).
\end{equation}

The key difference from our SG-MoE is that SCAR routes based on cross-attended features that already incorporate information from the target point cloud, while SG-MoE routes purely based on the query point cloud's semantic features.

As shown in Table~\ref{tab:ablation_additional}(middle), SCAR achieves 75.0\% RR, underperforming our vanilla SG-MoE by 1.3 points despite maintaining similar IR (44.4\% vs. 44.3\%).
This degradation suggests that incorporating inter-point-cloud relationships into routing introduces premature correspondence assumptions before the experts have processed the features.
The cross-attention operation in SCAR effectively performs a soft matching between source and target superpoints to guide routing, which may bias the routing decisions toward incorrect correspondences in challenging low-overlap scenarios.

In contrast, our SG-MoE design keeps routing decisions independent for each point cloud, based solely on its own semantic characteristics.
This allows experts to first specialize in processing semantic-geometric patterns within each point cloud, and then establish correspondences through the subsequent cross-attention layers where both semantic and refined geometric information are available.
Since correspondence information is already captured and refined in these subsequent cross-attention layers, introducing it prematurely in the routing stage provides no benefit and instead adds noise.
Our design of using only query-side semantic features for routing is simpler, more robust, and more effective.

\noindent\textbf{Number of Experts.}
We vary the number of experts $N_\text{expert}$ from 2 to 6 to study its impact on performance.
As shown in Table~\ref{tab:ablation_additional}(bottom), performance generally improves as we increase the number of experts from 2 to 4, with the best results at $N_\text{expert}=4$ (76.3\% RR).
Beyond 4 experts, performance degrades (75.5\% RR at $N_\text{expert}=5$, 75.7\% RR at $N_\text{expert}=6$), likely due to over-specialization and insufficient training samples per expert.

With only 2 experts (74.9\% RR), the model lacks sufficient capacity to capture diverse semantic-geometric patterns.
Three experts (75.1\% RR) provide some improvement but still underperform the optimal configuration.
The optimal choice of $N_\text{expert}=4$ strikes a balance between model capacity and training efficiency, allowing each expert to specialize in distinct semantic categories (e.g., planar structures like walls and floors, furniture, fixtures, and small objects) without over-fragmenting the feature space.

\noindent\textbf{Dependency on Semantic Quality.}
We evaluated SAGE's robustness by injecting three types of perturbations into semantic features on 3DLoMatch.

\begin{table}

\centering
\setlength{\tabcolsep}{4.2pt}
\begin{tabular}{l|ccccc}
\toprule
\textbf{Perturbation} & 10\% & 20\% & 30\% & 40\% & 50\%\\
\midrule
Gaussian noise      & 76.4 & 75.7 & 75.6 & 76.3 & 76.2  \\
Feature dropout     & 76.4 & 76.1 & 75.0 & 75.8 & 76.2  \\
Random replacement  & 75.8 & 75.1 & 74.8 & 74.8 & 73.5  \\
\bottomrule
\end{tabular}
\caption{Robustness to semantic feature degradation. Registration Recall (\%).}
\label{tab:semantics_dependency}
\end{table}

As shown in Table~\ref{tab:semantics_dependency}, SAGE maintains stable performance (RR \textgreater 75\%) under 50\% noise or dropout.
This robustness stems from the implicit regularization within our SG-MoE; through training, the router's MLP learns to bypass feature noise and prioritize stable semantic cues.

Specifically, while Gaussian noise and feature dropout represent unstructured degradations that the router can learn to filter as out-of-distribution noise, random replacement introduces structured, semantically plausible but incorrect features.
These ``adversarial'' semantic cues can mislead the router into assigning superpoints to experts specialized in irrelevant semantic contexts, providing them with a high but erroneous confidence. This structural confusion disrupts the correspondence search more severely than simple noise, as it actively guides the geometric refinement process toward semantically inconsistent regions, resulting in the observed gradual performance decline as the replacement ratio increases.

\begin{figure*}
  \centering
  \includegraphics[width=\linewidth]{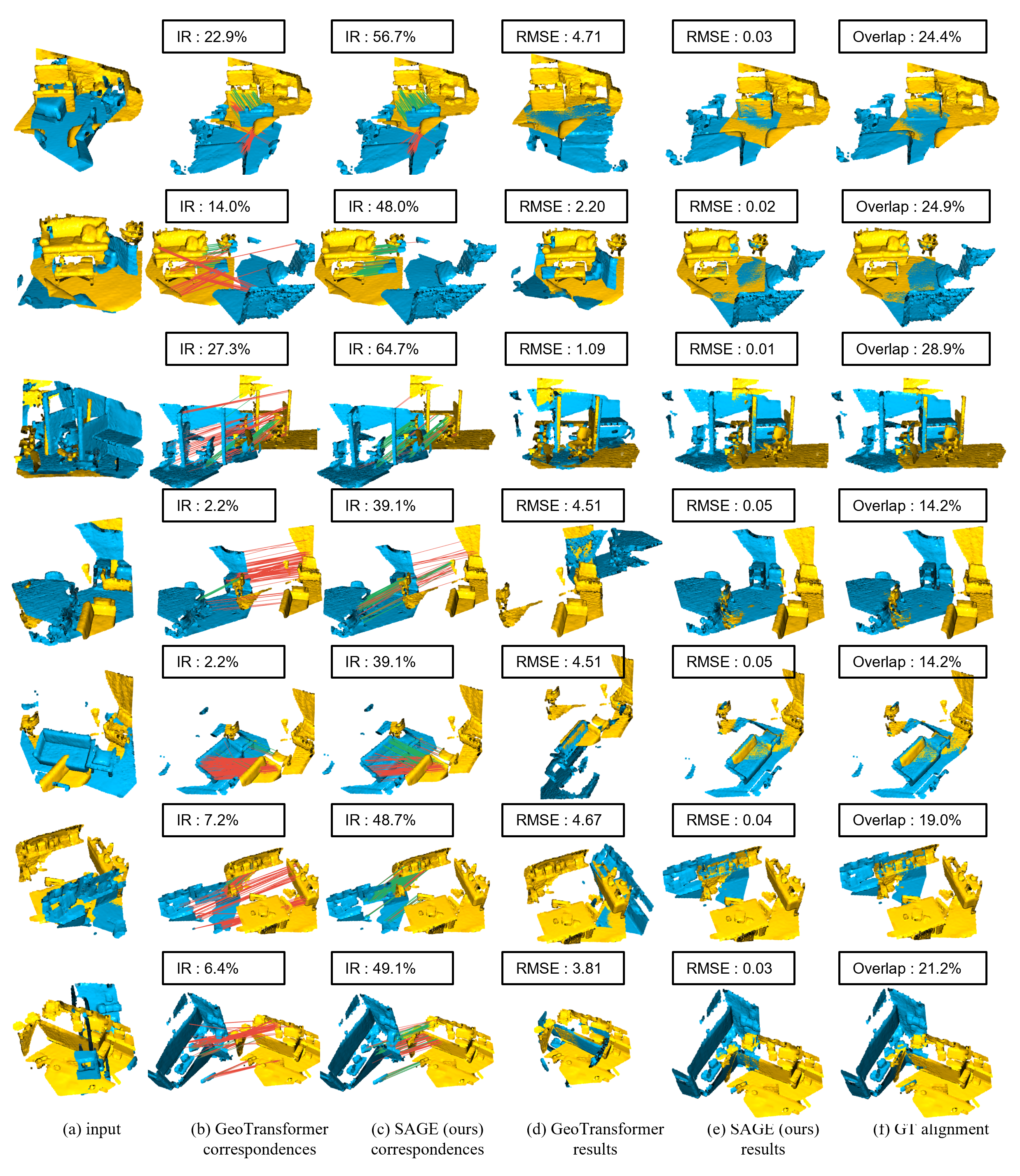}
  \caption{
  \textbf{Qualitative comparison on 3DLoMatch.}
  We compare the registration quality between GeoTransformer (baseline) and SAGE (ours) on challenging low-overlap pairs.
  From left to right:
  (a) Input source (yellow) and target (blue) point clouds.
  (b) Correspondences generated by GeoTransformer with IR.
  (c) Correspondences generated by SAGE with IR.
  (d) Registration result of GeoTransformer with RMSE.
  (e) Registration result of SAGE with RMSE.
  (f) Ground truth alignment with overlap ratio.
  SAGE consistently achieves accurate registration (low RMSE) even in cases where GeoTransformer completely fails (e.g., 3rd and 4th rows).
  }
  \label{fig:qualitative_sup}
\end{figure*}

\begin{figure*}
  \centering
  \includegraphics[width=\linewidth]{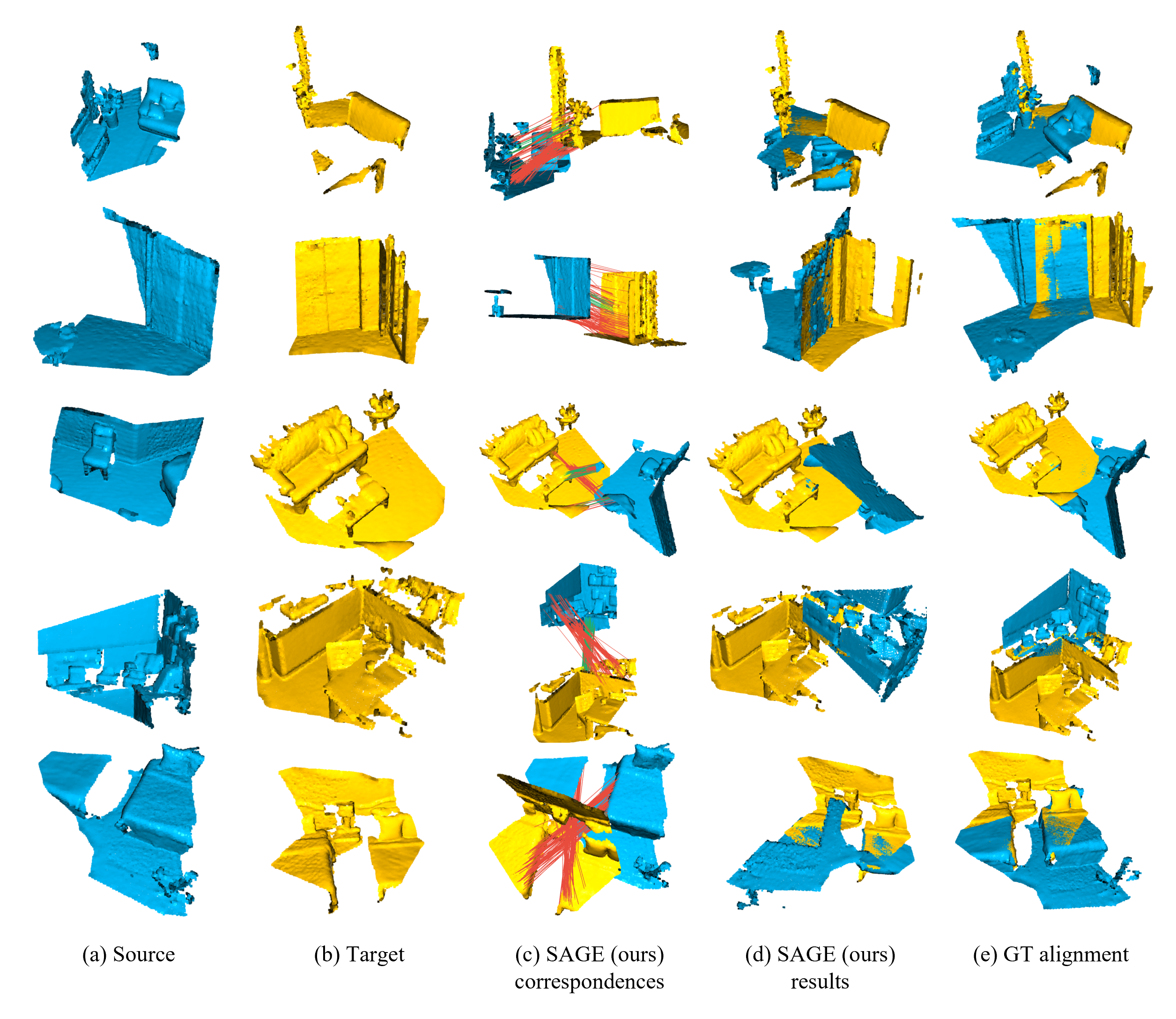}
  \caption{
  \textbf{Additional failure cases on 3DLoMatch.}
  Each row shows a challenging scenario where SAGE fails to achieve accurate registration.
  From left to right: (a) Source point cloud (cyan), (b) Target point cloud (yellow), (c) SAGE correspondences with Inlier Ratio (red lines: outliers, green lines: inliers), (d) SAGE registration result, (e) Ground truth alignment.
  These failures occur due to: (1) severe occlusion reducing the overlapping area, (2) geometrically indiscriminative regions (uniform walls, floors, ceilings), and (3) severe truncation of semantic regions, preventing the semantic encoder from extracting reliable semantic representations.
  When semantic structures are incomplete (e.g., only partial views of furniture), the frozen Sonata encoder cannot provide discriminative semantic features, as it was trained on complete object instances.
  }
  \label{fig:failure_supplementary}
\end{figure*}

\subsection{Qualitative Visualization}
\label{sec:d_4}
To explicitly demonstrate the robustness of SAGE in challenging scenarios, we provide qualitative comparisons against the baseline GeoTransformer on the 3DLoMatch benchmark.
Fig.~\ref{fig:qualitative_sup} visualizes the registration results on low-overlap pairs.

As observed in columns (b) and (c), GeoTransformer struggles to establish reliable correspondences in low-overlap regions, often resulting in low Inlier Ratios (IR) and incorrect registrations (high RMSE).
In contrast, SAGE successfully identifies correct correspondences (shown in green lines) by leveraging semantic cues from the Sonata foundation model, leading to accurate alignment even when the geometric overlap is limited (e.g., $<20\%$).
Specifically, SAGE effectively utilizes distinct semantic objects (e.g., furniture, fixtures) to anchor the registration, whereas the geometry-only baseline fails due to ambiguous geometric structures.

\subsection{Expert Routing Analysis}
\label{sec:d_7}
\begin{figure*}[t]
  \centering
  \includegraphics[width=0.7\linewidth]{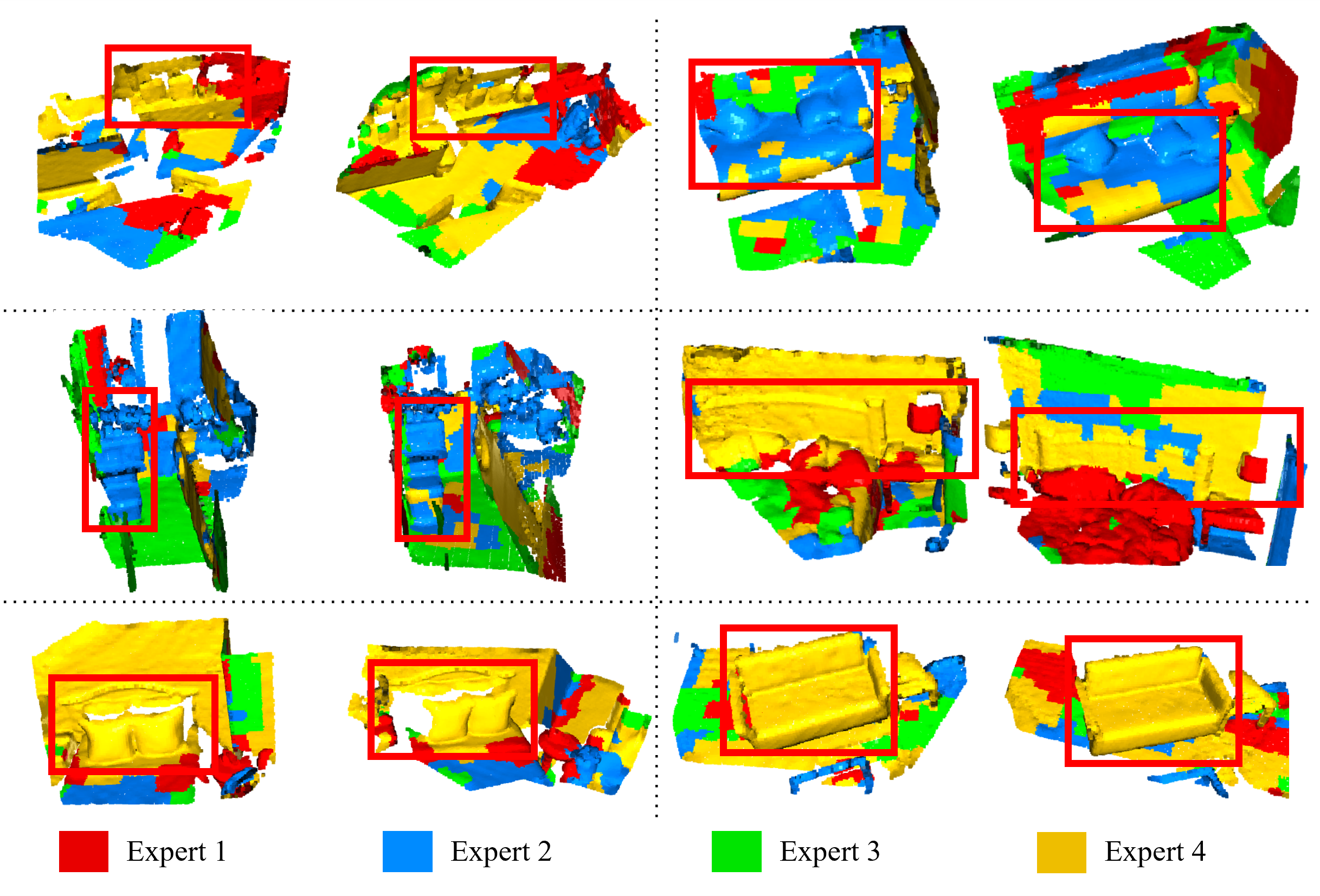}
  \caption{
  Expert routing patterns of SG-MoE across six scenes from 3DMatch and 3DLoMatch.
  Each scene is shown as a source-target pair, where superpoints are color-coded by their assigned expert (red, blue, green, yellow) and bounding boxes highlight semantically salient regions.
  The same semantic regions in the source and target point clouds are routed to the same expert, ensuring compatible feature representations for correspondence matching.
  This consistent functional specialization emerges across diverse scenes without explicit supervision, confirming that the frozen semantic features enable scene-agnostic and view-invariant expert routing.
  }
  \label{fig:cross_scene_expert}
\end{figure*}

To further analyze the semantic specialization of SG-MoE, we visualize the expert assignment maps across six diverse scenes sampled from both 3DMatch and 3DLoMatch in Fig.~\ref{fig:cross_scene_expert}.
Each superpoint is color-coded by its assigned expert (red, blue, green, yellow), and bounding boxes highlight semantically salient regions within each scene.

The visualization reveals that each expert develops a consistent functional role across all six scenes.
Expert~4 (yellow) consistently handles large, semantically prominent structures such as sofas, beds, and countertops, which serve as strong anchors for registration due to their distinctive geometry and semantic identity.
Expert~2 (blue) specializes in mid-scale objects such as chairs, shelves, and desks that provide complementary cues for correspondence establishment.
Expert~1 (red) is assigned to smaller, distinctive regions that are less dominant in terms of spatial extent but carry unique semantic signatures.
Expert~3 (green) primarily handles low-saliency background structures such as walls and floors, which are geometrically common across scenes and offer limited discriminative power for registration.

This consistent specialization pattern emerges without any explicit supervision on expert roles; it arises purely from the frozen Sonata semantic features that guide the routing decisions.
The fact that this functional division is preserved across scenes with substantially different layouts and object compositions demonstrates that SG-MoE routing generalizes beyond individual scene characteristics.

Importantly, within each scene pair, the source and target point clouds exhibit consistent expert assignments for the same semantic regions.
For instance, when a sofa appears in both the source and target scans, it is routed to Expert~4 (yellow) in both cases, even though the two scans are captured from different viewpoints and contain different degrees of occlusion.
This source-target consistency is essential for registration because the SG-MoE cross-attention layer processes the source and target features through their respective assigned experts.
When both sides of a correspondence are handled by the same expert, the resulting feature representations share a common specialized subspace, which directly improves the quality of correspondence matching.
Conversely, if source and target regions of the same object were routed to different experts, the resulting features would lie in incompatible subspaces, degrading the matching accuracy.
The observed consistency confirms that the frozen semantic representations provide a stable routing signal that is invariant to the viewpoint from which a scene is observed.

\section{Baseline Reproduction Details}
\label{sec:app_repro}
To make our comparison transparent and reproducible, this section describes how we obtain the baseline results in the main paper. All methods compared under the matched protocol (PEAL~\cite{yu2023peal}, PSReg~\cite{huang2025psreg}, and SAGE) are trained from scratch on the same NVIDIA H100 GPU and share the same training configuration as SAGE (Sec.~\ref{sec:b_4}): the Adam optimizer with an initial learning rate of $10^{-4}$ multiplied by $0.95$ after each epoch, weight decay $10^{-6}$, batch size $1$, and random seed $7351$. They also share the KPConv-FPN backbone (2.5\,cm initial voxel size, four stages), $128$ coarse targets, $256$ putative superpoint correspondences, and the geometric attention configuration ($4$ heads, $\sigma_d = 0.2$, $\sigma_a = 15^\circ$). Keeping these factors fixed isolates the effect of the routing design rather than confounding it with training or backbone differences.

\noindent\textbf{PEAL.}
The PEAL~\cite{yu2023peal} code is publicly released, so we build our PEAL baseline directly on the official repository. PEAL conditions its attention on an overlap prior. Although the prior data is released, the procedure used to generate it is not documented, which prevents us from reproducing it exactly. We therefore regenerate the overlap prior from the publicly released GeoTransformer~\cite{qin2022geotransformer} pre-trained model and train PEAL in our own environment under the shared configuration above. We use a budget of six refinement iterations, matching the iterative protocol used for SAGE and PSReg.

\noindent\textbf{PSReg.}
The official implementation of PSReg~\cite{huang2025psreg} is not publicly released, and several key hyperparameters, most notably the number of experts, are not reported in the paper. We therefore re-implement PSReg on the same framework as SAGE and train it from scratch. Following the design described in the PSReg paper, expert routing is driven by geometry-derived overlap information obtained from GeoTransformer~\cite{qin2022geotransformer}, with top-$1$ selection. To keep the comparison controlled, we set the number of experts to $4$, identical to SAGE; all remaining settings follow the shared configuration above.

\noindent\textbf{On the reproduced numbers.}
Our reproduced results for PEAL and PSReg fall below the values reported in their original papers. We attribute this to the absence of released pre-trained models, which forces retraining from scratch, and to the undocumented prior generation in PEAL and expert configuration in PSReg, which require our own implementation. Because these factors cannot be controlled by transcribing the originally reported numbers, we report all methods under a strictly matched protocol instead, so that the differences reflect the registration design rather than unspecified training conditions.

\noindent\textbf{Difference from the reproduced PSReg.}
SG-MoE (Sec.~\ref{sec:a_sgmoe}) and our PSReg re-implementation share the same backbone, attention, expert count ($4$), and top-$1$ routing; the only difference is the routing signal. PSReg fuses geometry-derived overlap information into the router input, whereas SG-MoE routes solely on the frozen semantic features while the experts still refine semantic-geometric features. The matched comparison in the main paper (Tables~1 and~2) therefore isolates the contribution of semantic-guided routing.

\section{Limitations}
\label{sec:app_limit}
While SAGE demonstrates significant improvements in low-overlap point cloud registration, several limitations warrant discussion for future work.

\noindent\textbf{Dependency on Pre-trained Semantic Models.}
SAGE relies on the quality and generalization capability of the frozen Sonata encoder.
The semantic degradation experiments in Table~\ref{tab:semantics_dependency} quantify this dependency: while SAGE is robust to unstructured noise (76.2\% RR even at 50\% Gaussian noise), random replacement of semantic features with plausible but incorrect ones causes a more pronounced decline to 73.5\% RR at 50\% replacement.
This indicates that the SG-MoE router is sensitive to structured semantic errors that mimic valid features, which could arise in practice when the encoder encounters object categories or scene types underrepresented in its training data (e.g., industrial equipment, medical instruments).
In such domains, Sonata may produce semantically ambiguous representations that mislead the router into suboptimal expert assignments.
As more diverse point cloud foundation models become available, replacing or complementing Sonata with encoders trained on broader data distributions could mitigate this limitation.

\noindent\textbf{Limited Effectiveness in Semantically Homogeneous Scenes.}
The scene-wise analysis in Table~\ref{tab:scene_results} reveals that SAGE's improvement over baselines is directly tied to the semantic diversity of each scene.
In the \textit{MIT Lab} scene, which consists of narrow corridors with repetitive walls and floors, SAGE (81.2\%) slightly underperforms GeoTransformer (83.9\%), and in \textit{Hotel 3}, all three methods achieve similar results (85.4--86.4\%).
The overlap-wise analysis in Table~\ref{tab:overlap_results} further corroborates this trend: at 50--60\% overlap, where geometric context is already sufficient, SAGE (98.4\%) is marginally below GeoTransformer (98.7\%).
These results indicate that when overlapping regions are dominated by uniform planar structures that lack semantic variation, the SG-MoE router cannot assign superpoints to meaningfully different experts, and the additional routing overhead provides no benefit.
Fig.~\ref{fig:failure_supplementary} illustrates concrete failure cases where this limitation manifests: severe occlusion leaves only featureless walls and floors in the overlapping area, and the truncation of semantic objects (e.g., only a partial view of a desk) prevents the frozen encoder from extracting discriminative representations.
We note that these scenarios are inherently challenging for all correspondence-based registration methods, as the lack of distinctive features is a fundamental obstacle that is not specific to semantic approaches.
Addressing robust registration in such low-texture, semantically homogeneous environments through temporal cues from sequential scans or multi-view constraints remains an important open problem for the community.

Despite these limitations, SAGE establishes a strong foundation for incorporating semantic understanding into point cloud registration, and the analyses presented in this work clarify the conditions under which semantic guidance is most and least effective, providing concrete guidance for future research in semantic-geometric fusion for 3D vision tasks.

\end{document}